\documentclass{WileyMSP-template}

\usepackage{amsmath}
\usepackage{amssymb}
\usepackage{ragged2e}
\usepackage{array}
\usepackage{booktabs}
\usepackage{stfloats}
\usepackage[table,xcdraw]{xcolor}
\usepackage{multirow}
\usepackage{pifont}
\usepackage{tikz}
\usepackage{makecell}
\usepackage{graphicx}
\usetikzlibrary{calc}
\usepackage{float}
\newcolumntype{Y}[1]{>{\centering\arraybackslash}m{#1}}
\newcolumntype{L}[1]{>{\raggedright\arraybackslash}p{#1}}
\newcolumntype{C}[1]{>{\centering\arraybackslash}p{#1}}
\newcommand{\cmark}{\ding{51}} 
\newcommand{\xmark}{\ding{55}} 

\begin{document}

\pagestyle{fancy}
\setlength{\headheight}{15pt}
\setlength{\footskip}{20pt}    

\newcommand{\moniesha}[1]{\textbf{\textcolor{red}{Moniesha: #1}}} 
\newcommand{\xing}[1]{\textbf{\textcolor{blue}{Xing: #1}}} 
\newcommand{\Damith}[1]{\textbf{\textcolor{pink}{Damith: #1}}} 
\newcommand{\David}[1]{\textbf{\textcolor{green}{David: #1}}} 

\title{A Real-World Grasping-in-Clutter Performance Evaluation \\
Benchmark for Robotic Food Waste Sorting}
\maketitle

\author{Moniesha Thilakarathna*, Xing Wang*, Min Wang, David Hinwood, Shuangzhe Liu, Damith Herath}


\begin{affiliations}
Moniesha Thilakarathna, Dr. Xing Wang, Dr. Min Wang, Dr. David Hinwood, Assoc Prof. Shuangzhe Liu, Prof. Damith Herath\\
Address: Faculty of Science and Technology, University of Canberra, 11 Kirinari St, Bruce ACT 2617, Australia
\\
Email Address: Moniesha.Thilakarathna@canberra.edu.au, Xing.Wang@canberra.edu.au

\end{affiliations}



\justifying

\vspace{-2pt}
\begin{abstract}

Food waste management is critical for sustainability, yet inorganic contaminants hinder recycling potential. Robotic automation presents a compelling approach by accelerating sorting through automated contaminant removal. Nevertheless, the diverse and unpredictable nature of contaminants introduces major challenges for reliable robotic grasping. 
Grasp performance benchmarking provides a rigorous methodology for evaluating these challenges in underexplored field contexts like food waste sorting. However, existing benchmarking approaches suffer from several limitations: over-reliance on limited simulation datasets and simplistic metrics such as success rate, inability to accommodate randomized grasping-in-clutter (GIC) environments, inability to account for object-related pre-grasp conditions, and a lack of comprehensive failure analysis.
To address these gaps, this work introduces GRAB, a comprehensive \textbf{G}rasping \textbf{R}eal-World \textbf{A}ttribute \textbf{B}enchmark, designed to evaluate GIC performance through: (1) diverse novel domain-specific deformable datasets, (2) advanced 6D grasp pose estimation, and (3) explicit evaluation of new pre-grasp conditions through graspability metrics. 
This benchmark systematically compares industrial grasping performance across three gripper modalities through experimental studies involving 1,750 grasp attempts across four high-fidelity randomized clutter levels, which enables a level of analysis rarely explored in prior work. 
The results reveal a clear hierarchy among graspability parameters, with object quality emerging as the dominant factor governing grasp performance across modalities, while pose quality and clutter exhibit comparatively moderate and context-dependent effects. 
Failure mode analysis shows that physical interaction constraints, rather than perception or control limitations, constitute the primary source of grasp failures in cluttered environments. By enabling the identification of dominant factors influencing grasp performance, the proposed benchmark provides a principled foundation for designing robust and adaptive grasping systems suited to complex, cluttered food waste sorting environments.

\end{abstract}


\vspace{-8pt}
\section{Introduction}\label{sec1}
\begin{figure*}[!t]
  \centering
  \includegraphics[width=0.9\textwidth,height=0.6\textheight,keepaspectratio]{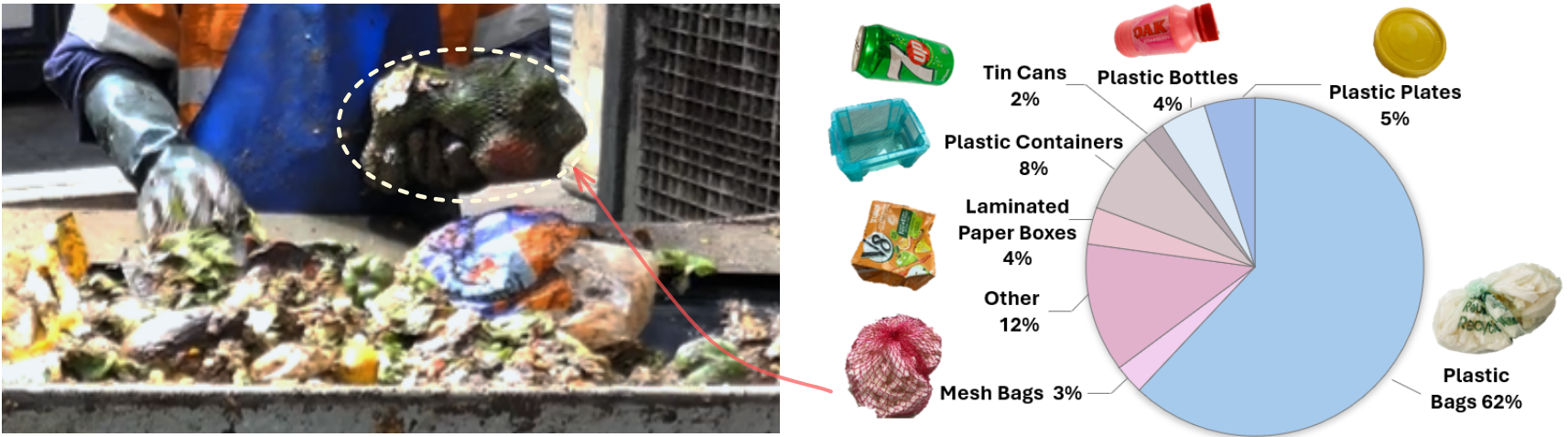}
  \vspace{-5pt}
  \caption{Industrial-scale variation of inorganic contaminants inside food waste}
  \vspace{-12pt}
  \label{wasteDistribution}
\end{figure*}

\vspace{-2pt}
The growing global population is compounding municipal solid waste (MSW) management challenges. Organic waste, comprising 19\% of MSW and primarily originating from food sources \cite{Pickin2022}, amounts to approximately 4.7 million tonnes in Australia. Recycling food waste is vital for a sustainable economy. However, the process is hindered by the presence of inorganic contaminants, as shown in Figure~\ref{wasteDistribution}. While uncontaminated food waste can be repurposed for low-emission applications such as protein farming \cite{NSW2}, segregation remains predominantly manual due to the heterogeneity of waste streams. This reliance on manual labour introduces challenges related to workforce shortages and worker discomfort. Consequently, the development of an efficient yet minimally viable automated sorting system presents a practical solution for improving the sustainability and scalability of food waste recycling processes.
\vspace{4pt}

However, robotic deployment in food waste sorting faces unique recognition and grasping challenges arising from cluttered environments \cite{Lin2023,lubongo2024} and the wide span of object variability caused by the coexistence of rigid and deformable contaminants \cite{Long2020}. The efficiency of robotic sorting systems can be enhanced through improvements in associated vision technologies or by augmenting the performance of the grasping system. While significant research has focused on advancing vision-in-clutter (VIC), there is a growing need to advance the physical performance capabilities of grasping-in-clutter (GIC), particularly in applications involving unpredictable clutter, like food waste sorting.
\vspace{4pt}

Benchmarking offers a rigorous, scientific methodology for evaluating the performance of a system and for determining optimal strategies for specific application domains. In the context of robotic sorting challenges, although multiple attempts have been made to benchmark VIC performance, efforts to benchmark the physical performance of GIC and to identify the optimal grasping modality for the problem domain remain limited, particularly for waste sorting applications, including complex domains such as food waste sorting. Not only in GIC but even in isolated object grasping scenarios, current grasp performance evaluation benchmarks primarily rely on metrics such as operational speed (mean picks per hour), most commonly grasp success rates  \cite{MahlerJ2018} drawn upon established object datasets such as YCB \cite{lerher2023}. However, the fixed taxonomies of these standardized datasets limit the adaptability of conventional evaluations to novel object categories in dynamic applications such as food waste sorting. Contemporary research underscores that grasp performance is highly dependent on application-specific conditions \cite{Fangerow2024, Qiu2023}, thereby necessitating grasp performance benchmarks that extend beyond simple success rate metrics. This recognition has spurred the development of advanced benchmarking protocols incorporating multi-object gripper-specific performance factors such as grasp strength and finger repeatability, and grasp efficiency using fingered grippers \cite{Chen2025, Falco2020}. Moreover, in real-world field contexts like food waste sorting, where random clutterness and a wide span of object variation are present, GIC performance cannot be measured solely by gripper performance metrics. This is because the properties of objects, their 6D placement within the surrounding environment, significantly affect physical performance in GIC. Therefore, comprehensive GIC performance evaluations must also account for environmental and object-related pre-grasp conditions \cite{Bottarel2020, Moniesha2025} and their relationship to grasp success or failure in benchmarking. 
\vspace{4pt}

We introduce GRAB (Grasping Real-World Attribute Benchmarking), a comprehensive benchmark to evaluate the grasping-in-clutter performance in real-world food waste contaminant sorting. It includes (1) diverse novel domain-specific deformable datasets, (2) advanced 6D grasp pose estimation, and (3) explicit evaluation of new pre-grasp conditions through graspability metrics. 
It is designed to systematically evaluate standard grasping modes and to assess their adaptability and limitations in robotic food waste sorting. 
Beyond conventional performance measurement metrics and limited datasets, GRAB proposes new object-related pre-grasp conditions and evaluates their relevance to the performance both in success and failure cases,  across a wide range of object variations and randomized cluttered environment setups. 

To summarise, our contributions are as follows:
\vspace{-5pt}
\begin{itemize}\setlength{\itemindent}{0pt}\setlength{\leftskip}{0pt}
    \item We introduce a comprehensive grasping-in-clutter (GIC) performance benchmark to evaluate the applicability of mainstream grasping modalities in extracting inorganic contaminants from food waste, through deliberation of object, pose and surrounding-related pre-grasp conditions to realize their impact on the physical performance of grasping modalities tested under a sensor-integrated real-world grasp benchmarking pipeline.
    \item We present a systematic and in-depth comparative performance evaluation of three uni-modal grippers across four high-fidelity cluttered scenes to identify how mainstream grasping modalities respond to diverse object categories in cluttered food waste scenes, providing valuable insights into the potential for cross-category grasping.
    \item We provide a comprehensive analysis of how pre-grasp conditions affect overall grasp performance of grasping modalities and their significant relationships.
    \item We present a comparative analysis of grasp failure modes and examine how pre-grasp conditions contribute to failure cases
\end{itemize}

\vspace{-10pt}
\section{Related Works}\label{sec2}

\vspace{-5pt}
\subsection{Robotic Sorting Towards Food Waste}

Traditional industrial sorting tasks typically involve handling uniform objects with predictable shapes, sizes, and positions, allowing for precise programming and stable grasp strategies \cite{li2019, zhang2019}. In contrast, waste streams contain a wide variety of objects with irregular geometries, varying material properties, and unpredictable spatial arrangements a key factor driving the growing interest in robotic waste sorting. Over the past decade,  the efficacy of robotic waste sorting has been substantially advanced through developments in artificial intelligence (AI) and automation, as demonstrated by industry leaders like AMP Robotics achieving up to 99\% sorting accuracy for recyclables using deep learning (DL) \cite{aschenbrenner2023}. This progress is further evidenced by recent innovations: Adetunji et al. developed a vacuum system employing convolutional neural networks to detect plastic bags \cite{Adetunji2025}, Maxence et al. created a 4-DoF (4D) parallel robot with a suction gripper for recyclables \cite{Leveziel2022}, and Xinxing et al. implemented a DL based mobile robot for construction waste \cite{Chen2022}. While robotic sorting typically targets recyclables such as glass, paper, plastic, and metals for resource recovery, a significant yet underexplored application is food waste sorting through robotic contaminant segregation, an area where robotic automation could enhance sorting efficiency and reduce worker health risks. Although it is significantly beneficial, the unpredictable nature of cluttered arrangements, along with the wide span of object variation hinder the applicability of conventional sorting procedures in this domain. This accentuates the need for a thorough evaluation of the pertinence of conventional grasping modalities to establish the basis for developing a minimum viable grasping solution for the targeted object categories in the application domain. 
\vspace{-8pt}

\subsection{Grasping Modalities for Cluttered Sorting}

A significant challenge in AI-driven robotic waste sorting involves developing grasping systems for extreme object diversity in highly cluttered environments \cite{Satav2023}. This demands enhanced physical intelligence of grasping modalities involved in the problem domain. Current industrial waste sorting systems primarily utilize two grasping modalities: suction grippers for lightweight objects with relatively flat surfaces \cite {aschenbrenner2023}, and rigid grippers for standard recyclables and heavier debris like construction waste \cite{Le2025, Xiao2020}. While effective within their respective niches, these traditional designs lack versatility across the full waste spectrum when compared to emerging soft grippers, which offer adaptive compliance particularly suited for handling deformable objects \cite{Chin2019}.  This diversity in uni-modal gripper capabilities has highlighted the need for universal or multi-modal grippers with integrated functionalities capable of handling a diverse spectrum of waste simultaneously, thereby increasing efficiency \cite{Satav2023}. For example, Rasoul et al. developed a vision-guided, tri-modal gripper that achieved 96\% success in controlled recyclable sorting \cite{Sadeghian2022}.  Despite significant progress in expanding the physical intelligence of grasping modalities from unimodal to multi-modal, extensive testing under cluttered scenarios remains lacking. Instead, these grippers are typically tested in isolated object-per-scene scenarios, which fail to account for the real-world complexity of GIC, where performance is challenged by the unpredictable nature of actual waste piles. Therefore, a thorough evaluation of the applicability of mainstream grasping modalities for a given problem domain must account for realistic cluttered scenarios rather than relying solely on testing with isolated object scenarios. 

\vspace{-8pt}

\subsection{Vision-in-Clutter for Waste Sorting}

Over the years, robotic vision has played a pivotal role in waste sorting by identifying regions of interest for grasping. For instance, machine learning algorithms like Faster R-CNN, Mask R-CNN, and YOLO enable waste identification through bounding boxes \cite{Wu2023, Zhang2022}. However, these 2D object identification approaches lack the spatial details required for successful grasping-in-clutter (GIC), necessitating algorithms that accurately identify the ideal grasp pose (GP) with position and orientation to inform object positioning in cluttered environments. Thus, early GP Detection (GPD) methods estimated 3D or 4D poses also as rectangles \cite{Joseph2014}, and this fixed-angle approach limited clutter adaptability \cite{Depierre2018}. Contemporary 6D-GPD addresses this by estimating full 6D configurations for enhanced flexibility \cite{Marcus2016}. Conventional 6-DoF grasp pose (GP) detection approaches typically rely on known geometric properties or precise 3D CAD models of objects. However, constructing accurate CAD models for every object present in a waste pile is unrealistic, which limits their applicability specially for applications such as food waste sorting. Algorithms that generate GPs directly from voxelized point clouds are able to overcome this challenge  \cite{Andreas2017,Liang2019}. Yet, a persistent challenge remains in robust estimation from noisy point clouds with limited datasets. GraspNet addresses this through synthetic-real data integration \cite{Fang2020}, a framework later advanced by GSNet's geometric structure encoder for improved reasoning in cluttered scenes \cite{Wang2021}. This lineage culminates in the AnyGrasp system, which builds upon GSNet to handle significant depth noise and incorporate stability awareness in GIC \cite{Fang2022}. 
\vspace{4pt}

\begin{table}[!b]
\centering
\vspace{-10pt}
\caption{Comparison of grasp performance benchmarking studies.}
\footnotesize
\renewcommand{\arraystretch}{1.15}
\setlength{\tabcolsep}{3pt}

\begin{tabular}{C{0.9cm} L{2.1cm} C{4.5cm} C{1.5cm} C{1.6cm} C{1.8cm} C{1.2cm} C{1.5cm} C{1.5cm}}
\toprule
\textbf{Study} & \multicolumn{2}{c}{\textbf{Object Data Set}} & \textbf{GIC} & \textbf{6D Grasp} & \multicolumn{3}{c}{\textbf{Object Related Pre-grasp conditions}} & \textbf{Failure} \\
\cmidrule{2-3}
\cmidrule{6-8}
\vspace{-4pt}
 & \textbf{Test Objects} & \textbf{Variability (\tiny Rigid to Deformable)} &  &  & \textbf{Property} & \textbf{Pose} & \textbf{Surrounding} & \textbf{Analysis}\\
\midrule

\cite{Falco2020}
& Artifacts
& \begin{tikzpicture}[x=1cm,y=0.5cm]
\draw (0,0) rectangle (3,0.3);
\fill[teal!70] (0,0) rectangle (0.5,0.3);
\end{tikzpicture} 
& \xmark & \xmark & \xmark & \xmark & \xmark & \xmark \\

\cite{Chen2025} 
& Convex shapes
& \begin{tikzpicture}[x=1cm,y=0.5cm]
\draw (0,0) rectangle (3,0.3);
\fill[teal!70] (0,0) rectangle (0.5,0.3);
\end{tikzpicture} 
& \raisebox{-1.2ex}{\makecell[c]{\cmark \\ \tiny(Pre-defined)}}
& \xmark & \xmark & \xmark & \cmark & \xmark\\

\cite{lerher2023}
& Modified YCB 
& \begin{tikzpicture}[x=1cm,y=0.5cm]
\draw (0,0) rectangle (3,0.3);
\fill[teal!70] (0,0) rectangle (3,0.3);
\end{tikzpicture} 
& \xmark & \xmark & \xmark & \xmark & \xmark & \xmark\\

\cite{Qiu2023}
& Food Items 
& \begin{tikzpicture}[x=1cm,y=0.5cm]
\draw (0,0) rectangle (3,0.3);
\fill[teal!70] (0,0) rectangle (2.5,0.3);
\end{tikzpicture} 
& \xmark & \xmark & \cmark & \xmark & \xmark & \cmark \\

\cite{Fangerow2024}
& Post consumer Plastics
& \begin{tikzpicture}[x=1cm,y=0.5cm]
\draw (0,0) rectangle (3,0.3);
\fill[teal!70] (0,0) rectangle (3,0.3);
\end{tikzpicture} 
& \raisebox{-1.2ex}{\makecell[c]{\cmark \\ \tiny(Pre-defined)}}
& \xmark & \cmark & \xmark & \cmark & \xmark\\

\cite{Bottarel2020}
& YCB 
& \begin{tikzpicture}[x=1cm,y=0.5cm]
\draw (0,0) rectangle (3,0.3);
\fill[teal!70] (0,0) rectangle (2,0.3);
\end{tikzpicture} 
& \raisebox{-1.2ex}{\makecell[c]{\cmark \\ \tiny(Pre-defined)}}
& \cmark & \xmark & \cmark & \cmark & \xmark\\

\cite{Avella2023}
& YCB + (18 objects) 
& \begin{tikzpicture}[x=1cm,y=0.5cm]
\draw (0,0) rectangle (3,0.3);
\fill[teal!70] (0,0) rectangle (3,0.3);
\end{tikzpicture} 
& \raisebox{-1.2ex}{\makecell[c]{\cmark \\ \tiny(Random)}}
& \cmark & \xmark & \xmark & \cmark & \xmark\\

\textbf{Ours}
&  \textbf{Real world contaminants}
& \begin{tikzpicture}[x=1cm,y=0.5cm]
\draw (0,0) rectangle (3,0.3);
\fill[teal!70] (0,0) rectangle (3,0.3);
\end{tikzpicture} 
& \raisebox{-1.2ex}{\makecell[c]{\cmark \\ \tiny\textbf{(Random)}}}
& \cmark & \cmark & \cmark & \cmark & \cmark\\

\bottomrule
\end{tabular}
\label{benchmarking}
\end{table}

The evolution of 6D grasp pose detection has expanded beyond parallel or pinch grasping scenarios and has also progressed toward suction-based grasping strategies. Dex-Net 3.0 emerged as a significant milestone in this era, generating suction poses from depth images, though its performance deteriorates in cluttered bins \cite{mahler2018}. In contrast, SuctionNet predicts stable poses directly from RGB-D images using end-to-end deep learning, eliminating heuristics for superior adaptability in waste environments \cite{cao2021}. These advancements in 6D grasp vision for cluttered sorting tasks provide strong evidence that complex waste sorting applications require accurate 6D grasp positioning to ensure safe grasp execution and achieve optimal physical performance of grasping modalities in grasping-in-clutter (GIC) scenarios. This requirement becomes particularly critical when evaluating the performance of different grasping modalities in grasp benchmarking studies that aim to identify which modality performs best for specific use cases. In such scenarios, the grasp vision system must operate at a high level of intelligence to reliably determine feasible grasp modalities for the application domain. However, current benchmarking approaches remain largely unexplored in evaluating the effectiveness of 6D grasp approaches for assessing the physical performance of grasping modalities within cluttered real world sorting applications.
\vspace{-5pt}

\subsection{Grasp Performance Benchmarking}

Grasp performance benchmarks characterize the elemental performance of robotic grasping modalities in a specific application domain and help to build a more complete picture of their overall performance in relevant applications. Grasp performance dependency on application-specific conditions \cite{MahlerJ2018} necessitate application-tailored benchmarking approaches \cite{Fangerow2024,Qiu2023} with performance evaluation metrics beyond success rate. However, enhancing the number of performance measures for evaluating gripper-specific factors through additional metrics such as grasp strength and finger repeatability alone is insufficient to fully address this need \cite{Falco2020}.  Since experimental design substantially influences grasping performance outcomes, it is important to include metrics relative to task complexity factors ranging from isolated single-object scenarios\cite{Falco2020,lerher2023, Qiu2023} to grasping in clutter (GIC) scenarios \cite{Chen2025,Fangerow2024,Bottarel2020} specially in evaluating grasp modalities for applications like waste sorting (Table \ref{benchmarking}). Yet, these evaluations are mostly conducted in predefined or customized clutter settings rather than accounting for the randomness that commonly appears in waste sorting scenarios.
\vspace{4pt}

On the other hand, representative object selection plays a decisive role when evaluating grasping modalities within a specific application domain. Nevertheless, most existing benchmarking procedures rely on standard datasets such as YCB objects or artifacts \cite{Falco2020,lerher2023,Chen2025,Bottarel2020}, which do not capture the full span of object variation from rigid to deformable that characterizes real-world waste object complexity. Furthermore, performance benchmarks that aim to identify optimal physical performance of grasping modalities must set other factors with high accuracy, particularly vision-in-clutter. However, existing benchmarking approaches have largely overlooked the role of 6D grasp positioning in their evaluation processes. 
\vspace{4pt}

Another critical consideration is that a successful grasp attempt requires prior understanding of whether the object to be grasped is graspable \cite{Fangerow2024}, especially in GIC. This includes gaining insights into object properties, identifying the correct 6D positioning of the object, and assessing the surrounding environmental complexity. However, current studies rely only on aggregated and occlusion-based metrics \cite{Avella2023} rather than decomposing these pre-grasp conditions and analysing it's impact on grasp performance. For example, even if an advanced 6D vision system accurately estimates the object pose, the gripper may still collide with the workspace during execution due to unexpected object deformation, resulting in grasp failure. This highlights the need for a decomposed analysis of pre-grasp conditions related to object properties, pose, and environment, enabling a more interpretable understanding of their individual contributions to grasp performance. Hence, effective grasp performance evaluation must explicitly consider these pre-grasp conditions to determine the suitability of grasping modalities for an application, a factor that remains largely overlooked in existing benchmarking approaches. 
\vspace{4pt}

Furthermore, these studies primarily rely on performance-oriented metrics and enable indirect failure diagnosis through decomposed scores such as success, stability, and obstacle avoidance, without explicitly identifying failure modes \cite{Bottarel2020}. At the same time, only a few studies attempt explicit failure categorization distinguishing between interaction, vision, and movement in grasping \cite{Qiu2023}, yet these remain limited in capturing the underlying mechanisms of failure, particularly in cluttered and complex interaction scenarios. 
\vspace{0.5pt}

In overall, existing research on grasp benchmarking lacks an integrated real-world performance benchmark that enables comprehensive evaluation of grasping modalities across a wide range of real-world objects, including highly deformable items, arranged in randomized cluttered scenarios, while jointly analysing performance metrics and decomposed pre-grasp conditions including object properties, 6D grasp positioning, and environmental factors and their influence on both grasp success and failure. 
\vspace{-8pt}

\section{Grasping-in-Clutter Performance Benchmark Methodology}\label{sec3}

This section details the methodology of the proposed benchmarking system, structured into four main components. First, we describe the foundational basis of the benchmark. Next, we present the grasping pipeline architecture, comprising the 6D grasp pose estimation algorithm, physical grasping modules, and path planning strategies. We then define the graspability parameters that characterize object-related pre-grasp conditions, along with the quantitative metrics used to evaluate the performance of grasping modalities in food waste sorting. Finally, we outline the benchmark execution protocol, including the experimental setup and procedures used to ensure a systematic and reproducible evaluation across randomized cluttered environments. 
\vspace{-8pt}

\subsection{Benchmarking Foundation}

\begin{figure}[t]
  \centering
  \vspace{-5pt}
  \includegraphics[width=\textwidth]{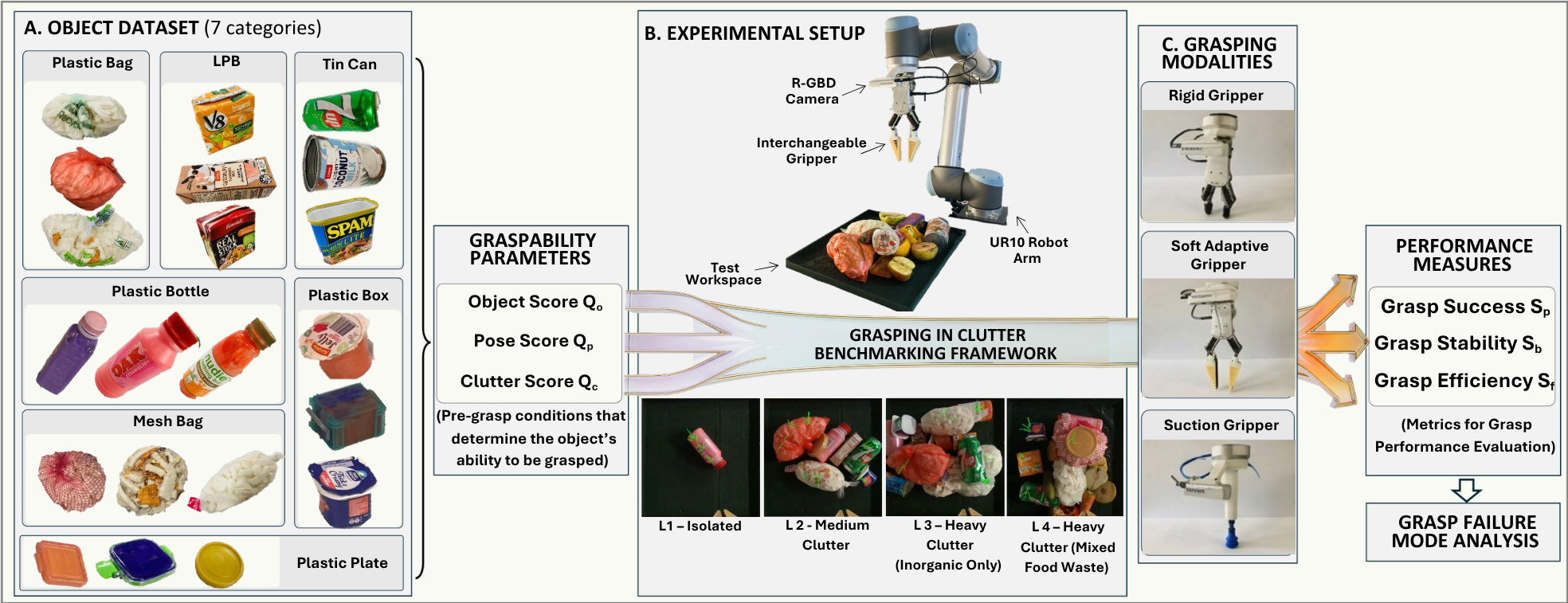}
  \caption{Overview of the proposed GRAB benchmark for grasping-in-clutter, illustrating the integration of (A) a real-world object dataset spanning seven categories, (B) an experimental robotic grasping setup with RGB-D perception and interchangeable grippers, (C) grasping modalities (rigid, soft adaptive, and suction), and (D) graspability parameters ($Q_o$, $Q_p$, $Q_c$) defining pre-grasp conditions. Performance is evaluated using success ($S_P$), stability ($S_B$), and efficiency ($S_F$) across multiple clutter levels (L1–L4), followed by failure mode analysis.}
  \vspace{-12pt}
  \label{conceptualization}
\end{figure}

The development of the benchmarking system (Figure \ref{conceptualization}) begins with the selection of representative inorganic object categories. The primary object dataset is derived from industrial field observations, which identified seven predominant categories of inorganic contaminants in food waste: Plastic Bags, Plastic Containers, Plastic Plates, Plastic Bottles, Laminated Paper Boxes (LPB), Mesh Bags, and Tin Cans (Figure \ref{wasteDistribution} and \ref{conceptualization}). These categories collectively account for approximately 88\% of inorganic materials present in food waste, forming the core testbed for the benchmark. To evaluate grasping performance, three mainstream grasping modalities are selected: a rigid parallel gripper, a soft adaptive gripper, and a suction gripper, as they reflect dominant industrial approaches in sorting applications. The experimental platform consists of a 6-DoF industrial robotic arm (UR10), an interchangeable gripper interface, an RGB-D camera for perception, and a dedicated workspace for object placement. Further details of the system architecture are provided in Section \ref{BenchPipeline}. 
\vspace{4pt}

\vspace{-5pt}

\vspace{4pt}
Within this benchmark, we define an object’s ability to be grasped (Graspability) in a cluttered environment primarily based on three pre-grasp conditions: the quality of the object itself, the quality of the candidate grasp poses estimated by 6D grasp pose detection, and the influence of clutter. Together, these conditions determine the overall status of the object before grasping. We identify these as graspability parameters: object score $Q_{O}$,  pose score $Q_{p}$ and clutter score $Q_{c}$ and they appear to be the key drivers that influence grasp performance.  To evaluate grasping performance, three complementary metrics are defined: grasp success, grasp stability, and grasp efficiency. The benchmark is executed through repeated trials across progressively increasing clutter levels, ranging from Level 1 (isolated objects) to Level 4 (heavily cluttered mixed food waste). This evaluation enables (i) identification of object-wise performance across different grippers, (ii) analysis of the influence of decomposed pre-grasp conditions on grasp performance outcomes, and (iii) systematic investigation of the root causes of grasp failures. 
\vspace{-8pt}

\subsection{Grasp Benchmarking Pipeline Architecture} \label{BenchPipeline}

\begin{figure}[t]
    \centering
    \vspace{-10pt}
    \includegraphics[width=\textwidth]{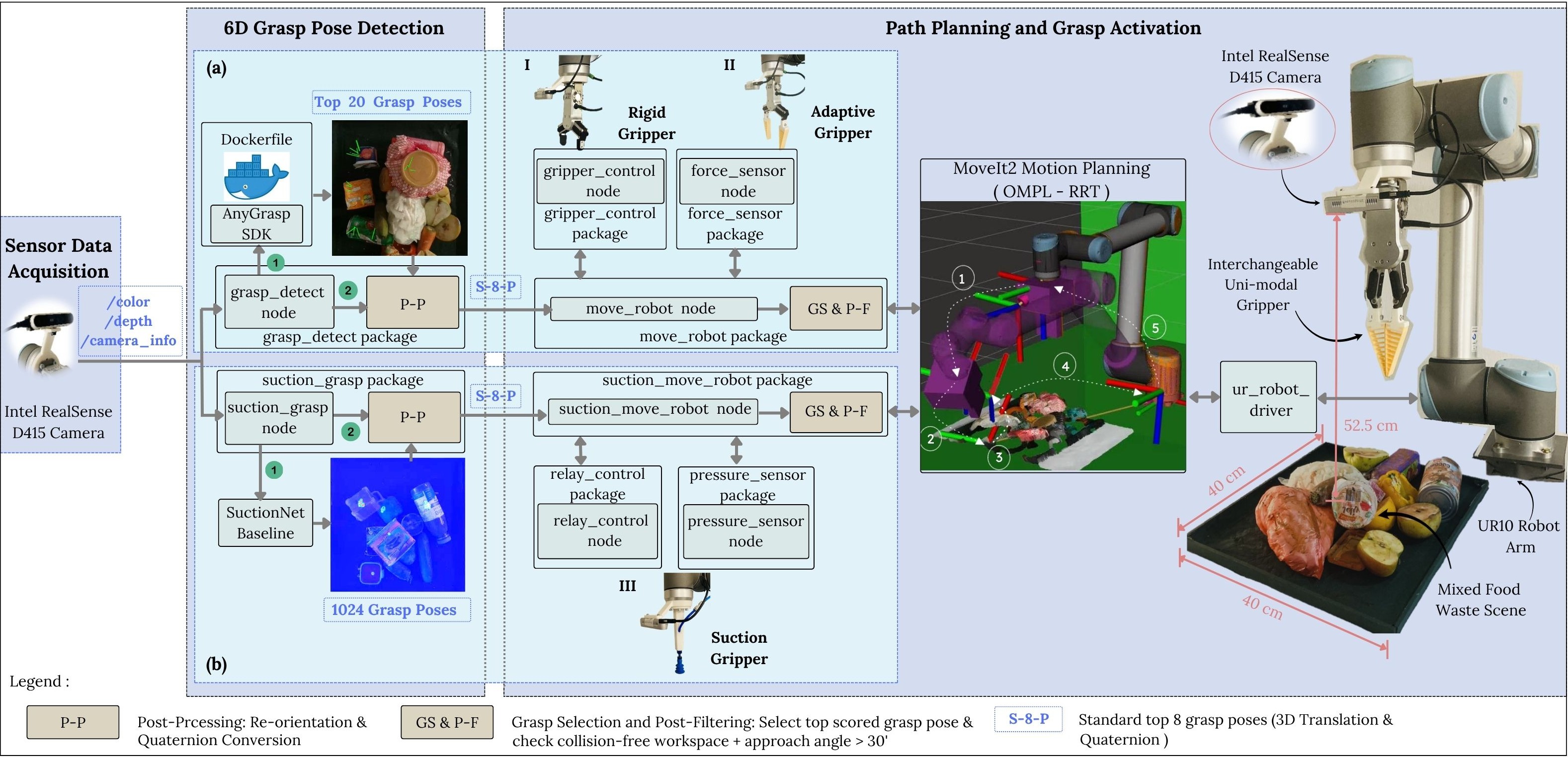} 
    \captionsetup{skip=2pt}
    \caption{Conceptualization of the real-world robotic grasping pipeline implemented using the ROS 2 control stack: (a) parallel grasping path for the rigid force-controlled parallel gripper (RG2) and adaptive Fin-Ray gripper, and (b) suction grasping path for the suction gripper. The pipeline illustrates RGB-D sensor data acquisition, 6D grasp pose detection (grasp generation using AnyGrasp and SuctionNet), post-processing, and selection of top-ranked feasible grasp poses, followed by path planning and grasp execution. The MoveIt 2 block illustrates the motion planning process, including the home pose, pre-grasp pose (purple arm), grasp pose within the waste scene, post-grasp pose, and release pose.}
    \vspace{-10pt}
    \label{Pipeline}
\end{figure}

The overall robotic grasping pipeline of the proposed benchmark (Figure \ref{Pipeline}) is structured around three core components: (i) a grasping system incorporating three distinct grasping modalities with integrated sensing capabilities,  (ii) a 6D grasp pose detection system built using advanced 6D grasp vision models, and (iii) a collision-free manipulation system for motion planning and execution. The robotic arm platform consists of an Intel RealSense D415 camera for 6D grasp pose detection, a 6-DoF UR10 robotic arm for manipulation and path planning, and a Lenovo Legion laptop equipped with an RTX 4070 GPU running Ubuntu 22.04 and ROS 2 Humble for system control and processing. 
\vspace{4pt}

The complete pipeline is orchestrated through ROS 2 packages, which manage sensor data acquisition, grasp pose generation and evaluation, grasp selection, motion planning, trajectory execution, and end-effector control. To accommodate the three grasping modalities, two specialized 6D grasp pose detection models are integrated into the pipeline. The AnyGrasp model \cite{Fang2022} is employed to generate high-quality candidate grasp poses for parallel grippers, owing to its robustness to sensor noise and generalization to unseen objects. For the suction gripper, SuctionNet \cite{cao2021}is utilized due to its capability to directly predict stable suction poses from RGB-D data. Both models are adapted and integrated within the ROS 2 control framework. Motion planning is performed using MoveIt 2 to ensure collision-free trajectories for the UR10 manipulator. End-effector actuation is managed through multiple interfaces: the RG2 gripper is controlled via the robot’s native controller, while the suction gripper is driven by a pneumatic system interfaced through an Arduino microcontroller integrated with ROS 2. 

\vspace{-7pt}
\subsubsection{Grasping Modalities}

\begin{figure}[!h]
    \centering
    \vspace{-6pt}
    \includegraphics[width=0.75\textwidth,height=0.30\textheight,keepaspectratio]{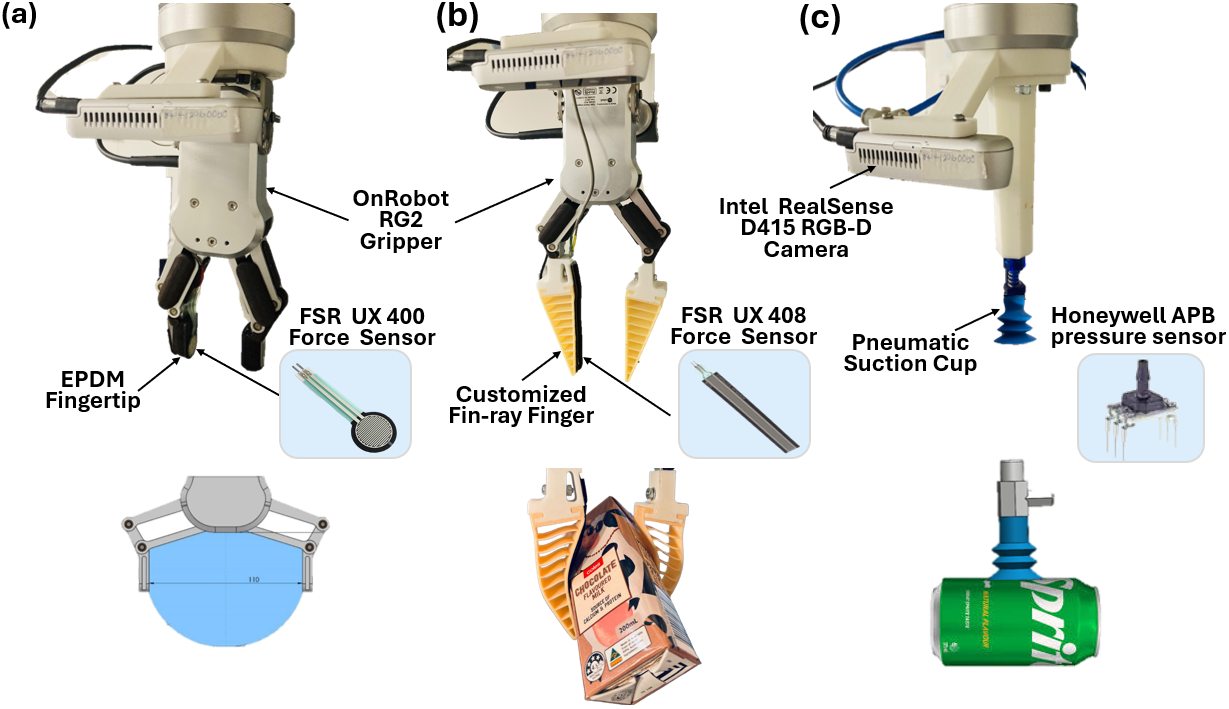} 
    \vspace{-8pt}
    \caption{ Three distinct grasping modes (a) Rigid force-controlled two-finger parallel gripper (RG2) with FSR UX 400 force sensor (b) Soft adaptive two-finger Fin-Ray gripper with FSR UX 408 force sensor (c) Suction gripper with Honeywell APB pressure sensor}
    \label{grippers}
\end{figure}

The rigid gripper used in this study is the industrial force-controlled parallel gripper OnRobot RG2 (Figure~\ref{grippers}).  In RG2, the parallel jaw mechanism ensures a solid and steady grip, rendering it especially appropriate for manipulating items with well-defined geometric characteristics. We set a constant force output of 40 N by controlling the driving motor's current output. 

The soft adaptive two-finger parallel gripper was developed by modifying the OnRobot RG2 gripper with baseline Fin-Ray soft fingers \cite{Wang2024}, which were specifically designed and 3D printed using ABS filament (1.75 mm) and integrated into the interchangeable finger slots of the RG2 The suction gripper was constructed using a bellows suction cup attached to a customized mounting mechanism. It produces optimal adhesion force at a maximum vacuum pressure of –80 kPa, where the suction cup adheres to the target surface by creating a pressure differential between the reduced pressure inside the suction cup and the ambient atmospheric pressure, thus generating a normal force that holds the object in place.

\vspace{-7pt}
\subsubsection{6D Grasp Pose Detection System} \label{subsubsecGPD}

The 6D vision system is initialized through an eye-in-hand calibration procedure to establish the transformation between the camera and robot coordinate frames. First, intrinsic camera parameters $(f_x, f_y, c_x, c_y)$ are estimated using a MATLAB calibration toolbox, achieving a mean reprojection error of 0.17 pixels. These parameters are used to project pixel coordinates $(u, v)$ with depth $Z_c$ into the camera frame $(X_c, Y_c, Z_c)$ as: 

\vspace{-20pt}
\begin{equation}
\begin{pmatrix}
X_c \\
Y_c \\
Z_c
\end{pmatrix}
=
\begin{pmatrix}
\frac{1}{f_x} & 0 & -\frac{c_x}{f_x} \\
0 & \frac{1}{f_y} & -\frac{c_y}{f_y} \\
0 & 0 & 1
\end{pmatrix}
\begin{pmatrix}
u \\
v \\
Z_c
\end{pmatrix}
\tag*{Equation (1)}
\end{equation}

\vspace{-4pt}
where $(c_x, c_y)$ denote the principal point in pixels, $(f_x, f_y)$ represent focal lengths, and $Z_c$ is the depth along the optical axis. 
\vspace{4pt}

To relate the camera frame to the robot base frame, eye-in-hand extrinsic calibration is performed to estimate the rigid transformation between the camera and the end-effector, represented by $\mathbf{T}_{\text{ee}}^{\text{cam}}$.  This transformation is combined with the forward kinematics of the UR10 manipulator to obtain the camera-to-base transformation ($\mathbf{T}_{\text{base}}^{\text{cam}}$) as given in below equation: 

\vspace{-6pt}
\begin{equation}
\mathbf{T}_{\text{world}}^{\text{cam}} = \mathbf{T}_{\text{base}}^{\text{cam}} = \mathbf{T}_{\text{base}}^{1} \mathbf{T}_{1}^{2} \mathbf{T}_{2}^{3} \mathbf{T}_{3}^{4} \mathbf{T}_{4}^{5} \mathbf{T}_{5}^{6} \mathbf{T}_{6}^{\text{ee}} \mathbf{T}_{\text{ee}}^{\text{cam}}
\label{eq:fk_chain}
\tag*{Equation (2)}
\end{equation}

where $\mathbf{T}_{\text{i}}^{\text{j}}$ denotes the homogeneous transformation from frame $i$ to frame $j$. 
\vspace{4pt}

After intrinsic and extrinsic eye-in-hand camera calibration, we proceeded to post-processing of the candidate grasp poses generated by the 6D vision models. From the grasp pose detection models, AnyGrasp used for parallel grasping, represents each grasp pose as: 

\vspace{-7pt}
\begin{equation}
    G = \begin{bmatrix} R & t & w \end{bmatrix}
    \tag*{Equation (3)}
\end{equation}

where $R \in \mathbb{R}^{3 \times 3}$ denotes proposed 6D orientation (rotation matrix), $t \in \mathbb{R}^{3 \times 1}$ represents the position (grasp center), and $w \in \mathbb{R}$ defines the gripper width (not part of the pose). Each grasp pose is associated with a quality score that evaluates grasp feasibility based on scene-level geometric reasoning and stability considerations.  Specifically, the model computes a grasp score by combining a geometric feasibility score with a stability-related term, where grasps that are likely to result in collisions are assigned a score of zero. For suction-based grasping, the SuctionNet model \cite{cao2021} represents each grasp candidate pose as: 

\vspace{-8pt}
\begin{equation}
    S = \begin{bmatrix} p & u \end{bmatrix}
    \tag*{Equation (4)}
\end{equation}
\vspace{-10pt}

where $p$ represents the suction contact point (position) and $u$ is the unit vector indicating the outward normal direction from the object surface. Similar to AnyGrasp, each suction candidate is associated with a quality score. SuctionNet processes RGB-D inputs to generate two complementary heatmaps: a seal score heatmap, which evaluates local surface suitability for forming an airtight seal, and a center score heatmap, which estimates the likelihood of each pixel serving as a stable suction grasp center. The final suction quality score is obtained by combining these two components, capturing both local geometric feasibility and global grasp stability. 
\vspace{4pt}

The outputs from both models provide the position and orientation of each candidate grasp pose. To make these representations suitable for path planning , all predicted grasp poses are post-processed. Since the predicted orientations are defined in the models’ reference frames, they are re-oriented to align with the end-effector coordinate convention of the robot, specifically by matching the predicted approach direction with the gripper’s z-axis. The resulting orientations are then converted into quaternion representations, producing standardized 6D poses (3D translation and quaternion) accompanied by the corresponding model-generated grasp scores. 

\subsubsection{Path Planning and Grasp Activation} \label{subsubsecplanning}

Path planning is initiated by selecting feasible grasp poses from the post-processed candidate set generated by the 6D grasp pose detection models. From these candidates, the top eight grasp poses are selected based on their associated model-generated grasp scores. Before executing motion planning, these selected grasp poses, initially defined in the camera coordinate frame ($\mathbf{T}_{\text{cam}}^{\text{pose}}$), are transformed into the robot base (world) coordinate frame using the calibrated camera-to-base transformation:

\vspace{-6pt}
\begin{equation}
\mathbf{T}_{\text{world}}^{\text{pose}} = \mathbf{T}_{\text{world}}^{\text{cam}}  
\mathbf{T}_{\text{cam}}^{\text{pose}}
\tag*{Equation (5)}
\end{equation}

where $\mathbf{T}_{\text{world}}^{\text{cam}}$ is obtained from the calibration process described in Equation (2).
\vspace{4pt}

\begin{figure}[hbt!]
  \centering
  \vspace{-10pt}
  \includegraphics[width=0.44\textwidth,height=0.60\textheight,keepaspectratio]{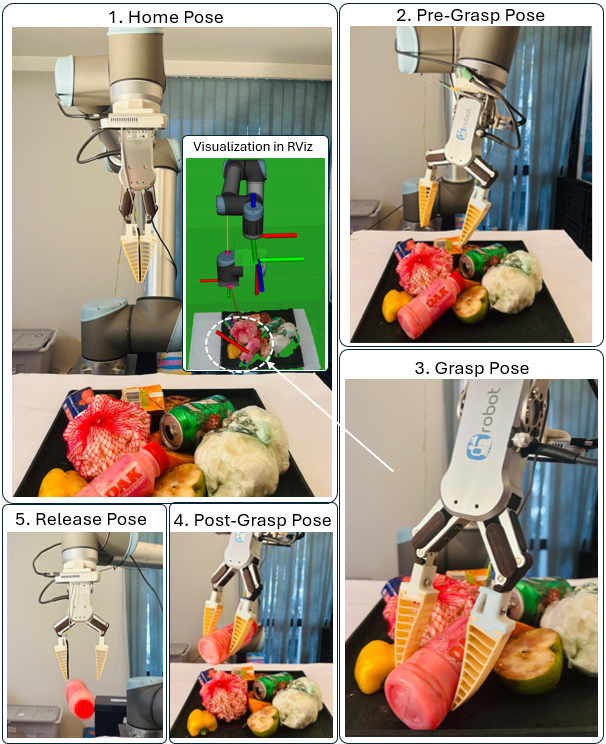}
  \caption{Key stages of the grasp execution process: (1) home pose with grasp pose visualization in RViz, (2) pre-grasp pose, (3) grasp pose during object engagement, (4) post-grasp pose after successful object lifting, and (5) release pose.}
  \vspace{-10pt}
  \label{graspcycle}
\end{figure}
\vspace{4pt}

Following this transformation, a final executable grasp pose is selected through a post-filtering process based on two criteria: (1) the pose must be collision-free with respect to the test workspace, verified using the MoveIt Flexible Collision Library, and (2) the approach vector must form an angle of at least 30 degrees with the test workspace plane to reduce occlusions and ensure safe execution. Once a valid grasp pose is identified, a sequence of five key poses is generated for execution, including the pre-grasp pose, grasp pose, post-grasp pose, and release pose  (Figure \ref{Pipeline} and \ref{graspcycle}). The Pre-Grasp pose is calculated by offsetting backwards along the approach vector, facilitating seamless transitions and avoiding collisions. The integration of both pre-grasp and post-grasp positions is essential for preventing unintended collisions with adjacent objects during execution.
\vspace{4pt}

MoveIt2 serves as the principal framework for motion planning, utilizing the Rapidly-exploring Random Tree (RRT) - state-of-the-art route planning algorithm from the Open Motion Planning Library (OMPL) to produce collision-free trajectories. To improve trajectory accuracy, Cartesian path planning utilizing linear interpolation is employed, ensuring smooth and predictable end-effector movement. This is further refined by enforcing velocity and acceleration constraints to guarantee kinematically feasible movements that prevent sudden jerks and protect grasp integrity.
\vspace{-8pt}

\subsection{Graspability Parameters: Pre-Grasp Conditions}

In order to ensure fair evaluation of each grasping mode, the object being grasped must itself demonstrate a reasonable ability to be grasped by the gripper. In this benchmark, each grasp cycle is quantitatively analysed through three graspability parameters: object score $Q_{o}$, pose score $Q_{p}$, and clutter score $Q_{c}$.

\vspace{-6pt}
\subsubsection{Object  Score ($Q_{o}$)}

In the proposed benchmark, the object score is associated with the intrinsic characteristics of each individual object; therefore, the score is assigned to each object prior to grasp trial execution. We collected three representative items from each of the seven categories, resulting in a total of 21 distinct objects as the object dataset for experimentation. All selected objects weighed less than 100g, which is below the allowable payload of the grippers, and had a cubic volume within 20 × 10 × 10 cm³, making them suitable for the gripper widths and the negative pressure of the suction gripper. Unlike published datasets, this object dataset includes both rigid and deformable objects, covering a significant span of geometric variation. Since weight and size were controlled to satisfy the gripper constraints, object deformation or shape variation emerged as the primary object-related factor influencing the ability to be grasped. Therefore, the deformation was parameterized as the object-related graspability condition, defined as the Object Score $Q_{o}$.
\vspace{4pt}

\begin{figure}[!b]
  \centering
  \vspace{-10pt}
  \includegraphics[width=0.75\textwidth,height=0.38\textheight,keepaspectratio]{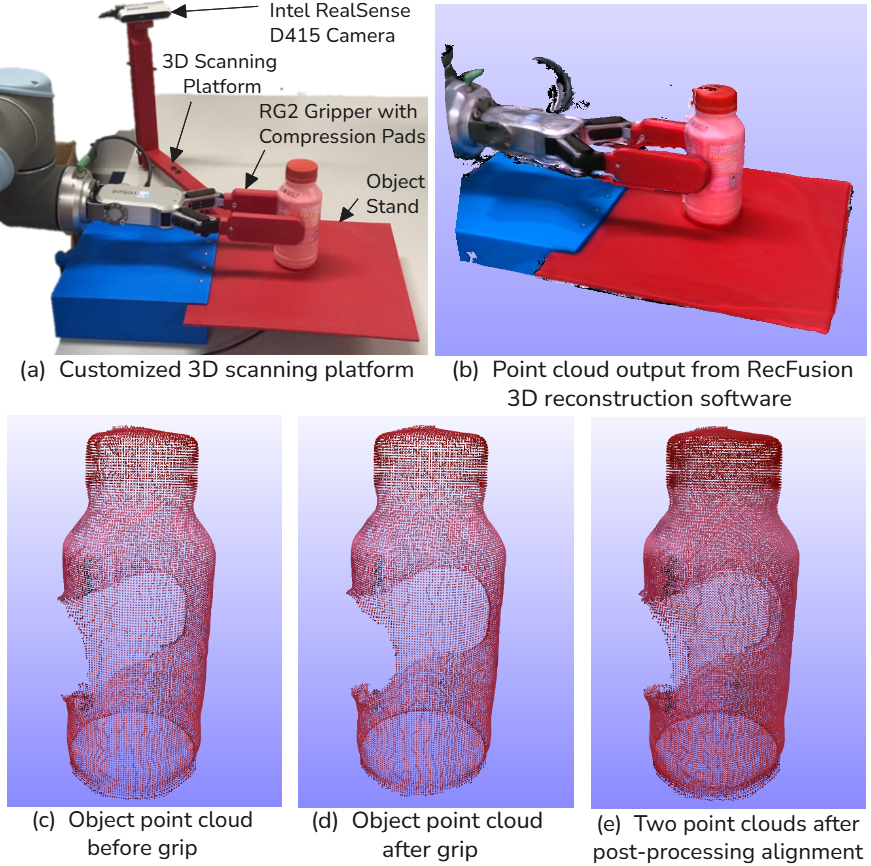}
  \caption{Overview of 3D scanning and point-cloud processing workflow for DCD computation (a) Customized 3D scanning and compression platform, (b) RecFusion 3D reconstruction, (c) point cloud of the undeformed object before gripping, (d) point cloud of the deformed object under compression, and (e) aligned point clouds illustrating geometric differences due to deformation.}
  \vspace{-12pt}
  \label{dcdscanning}
\end{figure}

The primary methodology for obtaining the object deformation value involved measuring the Density-Aware Chamfer Distance (DCD) \cite{Wu2021} between two 3D point clouds of the object: one captured before grasping and the other after grasping under an applied compression force of 40 N. DCD quantifies the geometric discrepancy between the two point clouds by computing the average nearest-neighbour distance while incorporating local point density information, thereby providing a more balanced and representative measure of shape variation. In the context of deformation analysis, this enables the capture of subtle geometric changes in object structure between pre- and post-grasp states. Moreover, DCD offers a significant advantage over standard metrics such as Chamfer Distance or Earth Mover’s Distance, as it is robust to uneven point density and the presence of outlier points \cite{Greenland2025}. This robustness ensures consistent and reliable measurement even under partial occlusions, making DCD particularly well-suited for deformation analysis using incomplete point clouds. The DCD algorithm is defined as:

\vspace{-12pt}
\begin{align}
d_{\mathrm{DCD}}(S_{1}, S_{2}) = \tfrac{1}{2} \Bigg( 
& \frac{1}{|S_{1}|} \sum_{x \in S_{1}} 
\left( 1 - \frac{1}{\hat{n}_{y}} e^{-\alpha \lVert x - \hat{y} \rVert_{2}} \right) \nonumber \\[6pt]
+ \; & \frac{1}{|S_{2}|} \sum_{y \in S_{2}} 
\left( 1 - \frac{1}{\hat{n}_{x}} e^{-\alpha \lVert y - \hat{x} \rVert_{2}} \right) 
\Bigg)
\tag*{Equation (6)}
\end{align}
\vspace{-6pt}

Here, $\hat{y} = \min_{y \in S_{2}} \lVert x - y \rVert_{2}$ and $\hat{x} = \min_{x \in S_{1}} \lVert y - x \rVert_{2}$ denote the nearest neighbours of points $x$ and $y$, respectively. The metric computes the average discrepancy between each point in one point cloud ($S_{1}$) and its closest point in the other ($S_{2}$), while incorporating local density normalization through $\hat{n}_{x}$ and $\hat{n}_{y}$. The exponential term maps Euclidean distances into a bounded range, ensuring that the metric remains within $[0,1]$. In practice, this formulation approximates small-distance behaviour using a first-order Taylor expansion, such that smaller distances are mapped close to 0 while larger distances progressively saturate towards 1. The scalar parameter $\alpha$ controls the sensitivity of this mapping, governing how rapidly the metric penalizes deviations and, consequently, the degree of deformation.
\vspace{4pt}

We developed a customized experimental platform to capture 3D point clouds of objects before and after being gripped under a 40 N compression force (Figure \ref{dcdscanning} (a) and (b)) along both the x and y axes. The setup consisted of a UR10 robot arm, an RG2 gripper with custom-made compression pads, and a customized 3D scanning platform based on RecFusion software integrated with an Intel RealSense D415 RGB-D camera.  RecFusion generates point clouds by projecting RGB-D data into 3D space using camera intrinsics, followed by frame-to-frame registration and volumetric TSDF fusion to produce a dense and globally consistent 3D reconstruction.  To evaluate these point clouds, the grasped objects needed to be segmented from the 3D scene and isolated from both the grippers and the background. To facilitate this process, the scene was configured to have high contrast: the test objects were placed on a plain background, and the test gripper compression pads were printed in the same colour. Background removal and filtering were performed using both RecFusion and MeshLab software. First, a spatial cropping operation was applied to retain only the region of interest containing the object. Subsequently, color-based thresholding was used to separate the object from the background.
\vspace{4pt}

\begin{figure}[!h]
  \centering
  \vspace{-10pt}
  \includegraphics[width=0.60\textwidth,height=0.30\textheight,keepaspectratio]{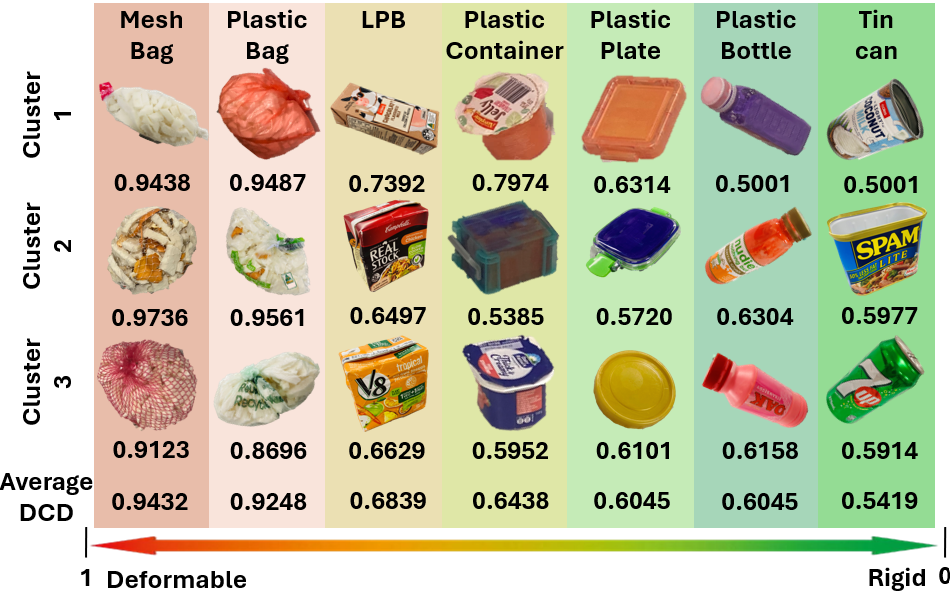}
  \caption{Object dataset with their corresponding DCD values}
  \vspace{-10pt}
  \label{dcd}
\end{figure}

To accurately measure deformation and correct for small in-hand orientation shifts, the pre- and post-grasp point clouds were aligned during post-processing (Figure  \ref{dcdscanning} (c)-(e)). Since different parts of the object become visible before and after grasping, the two point clouds exhibit varying feature sets, making alignment essential. The alignment was performed in two stages: first, the principal axes of the point clouds were aligned using MeshLab to provide an initial coarse registration, then, the Iterative Closest Point (ICP)\cite{Wang2017} algorithm was applied to refine the alignment and minimize the residual distance between the two point clouds. Based on the calculated DCD values for x and y planes point cloud pairs captured before and after grasping, the averaged DCD value was computed to obtain a representative deformation measure for each object.  Higher values of $Q_{o}$ indicate increased deformation, while lower values represent more rigid objects.  The finalized object scores are presented in Figure  \ref{dcd}.

\vspace{-6pt}
\subsubsection{Pose Score ($Q_{p}$)}

The pose score $Q_{P}$ quantifies the quality of candidate grasp poses estimated by the 6D grasp pose detection system, representing the 6D positioning of the object in cluttered scenes. In this work, $Q_{P}$ is derived from the grasp scores generated by the respective models. As outlined in Section \ref{subsubsecplanning}, for each grasp attempt, the top-ranked grasp poses proposed by the 6D grasp detection system are further filtered through post-processing criteria to select the highest-ranked valid grasp pose with execution feasibility. The quality score of this selected grasp pose is taken as the primary pose score $s$ for a grasp attempt.  Since AnyGrasp \cite{Fang2022} and SuctionNet \cite{cao2021} produce scores with different physical interpretations: geometric grasp stability and suction seal feasibility, it is required to interpret these scores within the context of their respective grasping modalities rather than directly comparing their absolute values. Specifically, for a given grasp attempt, the selected grasp quality score $s$  is normalized within the corresponding model-specific score range using min–max normalization and assigned as $Q_{P}$:

\vspace{-8pt}
\begin{equation}
Q_{P} = \frac{s - s_{\min}}{s_{\max} - s_{\min}}, \quad s \in \{S^{AG}, S^{SN}\}
\tag*{Equation (7)}
\end{equation}
\vspace{-8pt}

where $S^{AG}$ and $S^{SN}$ denote the grasp scores generated by AnyGrasp and SuctionNet, respectively, and $s_{\min}$ and $s_{\max}$ represent the minimum and maximum scores within the corresponding model’s grasp candidate set. Higher values of $Q_{P}$ indicate more reliable and executable grasp poses within the context of the corresponding grasping modality. If no valid grasp pose is generated for a given object, the pose score is assigned a value of $0$. Grasp execution and subsequent performance evaluation are conducted only when valid grasp poses are successfully identified. It is important to note that $Q_{P}$ reflects the relative ranking of grasp candidates within each model rather than an absolute measure of grasp quality across different grasping modalities. This formulation preserves the intrinsic characteristics of each model’s scoring mechanism while avoiding direct comparison between heterogeneous grasp representations. Consequently, the 6D grasp pose detection system is treated as a fixed component, enabling the evaluation to isolate the physical grasping capability of the grasping modalities as the primary variable under investigation.

\vspace{-6pt}
\subsubsection{Clutter Score ($Q_{c}$)}

The third graspability condition involves quantifying the impact of the surrounding environment when an object is located within a cluttered scene. The clutter score $Q_{C}$ in $[0,1]$ captures global environmental complexity by characterizing the object density within the scene. The proposed clutter score is defined as a global clutter-density descriptor based on two parameters: the proportion of the workspace occupied by objects (occupancy) and the total number of objects present in the scene at a given time.
\vspace{4pt}

\begin{figure*}[!h]
    \centering
    \includegraphics[width=\textwidth]{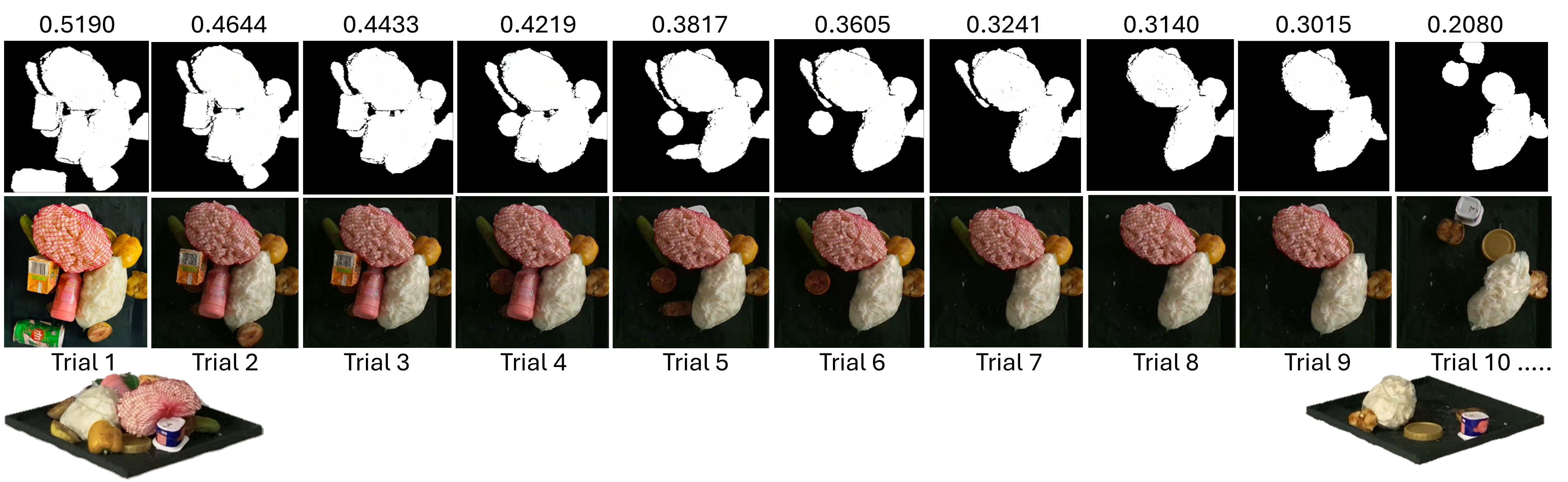} 
    \captionsetup{skip=2pt}
    \vspace{-30pt}
    \caption{Variation of occupancy during a grasp cycle for an object cluster}
    \label{Occupancy}
\end{figure*}

This formulation is designed to provide a consistent and controlled representation of clutter progression across successive grasp cycles.  Each grasping run begins with a larger number of objects, resulting in higher occupancy, and as objects are grasped and removed, both the number of objects and the occupancy gradually decrease  (Figure  \ref{Occupancy}). In this framework, clutter is treated as a global environmental variable that reflects the overall interaction constraints imposed on the gripper, while avoiding sensitivity to object-specific geometric variations that may introduce uncontrolled variability in clutter quantification. The clutter score is computed only for trials where more than one object exists in the scene.  In single-object trials, where no additional objects are present in the scene, clutter impact is not applicable.
\vspace{-10pt}

\begin{equation}
Q_{c} =
\begin{cases}
N \times O, & N_{\text{initial scene}} > 1\\[6pt]
0, & N_{\text{initial scene}} = 1 \quad \text{(no clutter)} 
\end{cases}
\tag*{Equation (8)}
\end{equation}

\vspace{-10pt}
\begin{equation}
N = \frac{N_{\text{before trial}}}{N_{\text{initial scene}}}, \qquad
O = \frac{O_{\text{before trial}}}{O_{\text{initial scene}}}
\quad \text{where } O \text{: occupancy ratio and } N \text{: object ratio.}
\tag*{Equation (9)}
\end{equation}

Occupancy was calculated through a sequence of filtering steps applied to the RGB-D data. The workspace mask was first binarised to separate the valid workspace from the background. Contours were then extracted to retain only the outermost boundary. The contour was compressed by removing redundant points and retaining only the essential vertices \cite{Yang2024}. The largest contour was selected as the effective workspace, acting as a contour-based spatial filter, and further refined by horizontal and vertical trimming to eliminate boundary artifacts. In addition, a depth filter was applied to the depth image, restricting valid object points to a predefined range of 250–525 mm to exclude noise and background pixels outside the graspable region. Finally, occupancy was computed as the ratio of object pixels (passing all filters) to the total workspace pixels, expressed as a percentage.

\vspace{-6pt}
\subsection{Performance Evaluation Metrics}

While graspability parameters characterize the pre-grasp conditions influencing an object’s ability to be grasped, grasp performance metrics quantify the physical performance of the grasping modalities during grasp execution. In this benchmark, grasp performance is evaluated using three complementary metrics: grasp success $S_{p}$, grasp stability $S_{b}$, and grasp efficiency $S_{f}$.  

\vspace{-6pt}
\subsubsection{Grasp Success Score ($S_{p}$)}

Grasp success score $S_{p}$ represents the outcome of a grasp attempt in terms of successful object acquisition, indicating whether the object is effectively grasped at the contact stage. It is defined as a discrete metric capturing grasp execution: 
\vspace{-5pt}
\[
S_{p} =
\begin{cases}
1, & \parbox[t]{.7\linewidth}{if the object is grasped and successfully moved to the pre-grasp pose} \\ 
0.5, & \parbox[t]{.7\linewidth}{if the object is grasped but dropped while moving to the pre-grasp pose} \\ 
0, & \text{if the grasp attempt fails}.
\end{cases}
\tag*{Equation (10)}
\]

\vspace{-8pt}
\subsubsection{Grasp Stability Score ($S_{b}$)}

The Stability score $S_{b}$ measures how stable the grasp remains during a successful execution. It reflects the gripper’s ability to securely hold the object while moving it from the grasp pose to the drop pose. To evaluate this, we implemented an object-holding tracking (OHT) system  for all three grippers. For the rigid and Fin-Ray grippers, force sensors (FSR UX-400 and FSR UX-408, respectively) were embedded and calibrated accordingly (Figure \ref{grippers}). The sensor actuation circuitry was controlled via a relay module integrated with an Arduino microcontroller, with communication managed through ROS 2 (Figure \ref{Pipeline}).
\vspace{4pt}

\begin{figure}[!t]
  \centering
  \vspace{-10pt}
  \includegraphics[width=0.50\textwidth,height=0.3\textheight,keepaspectratio]{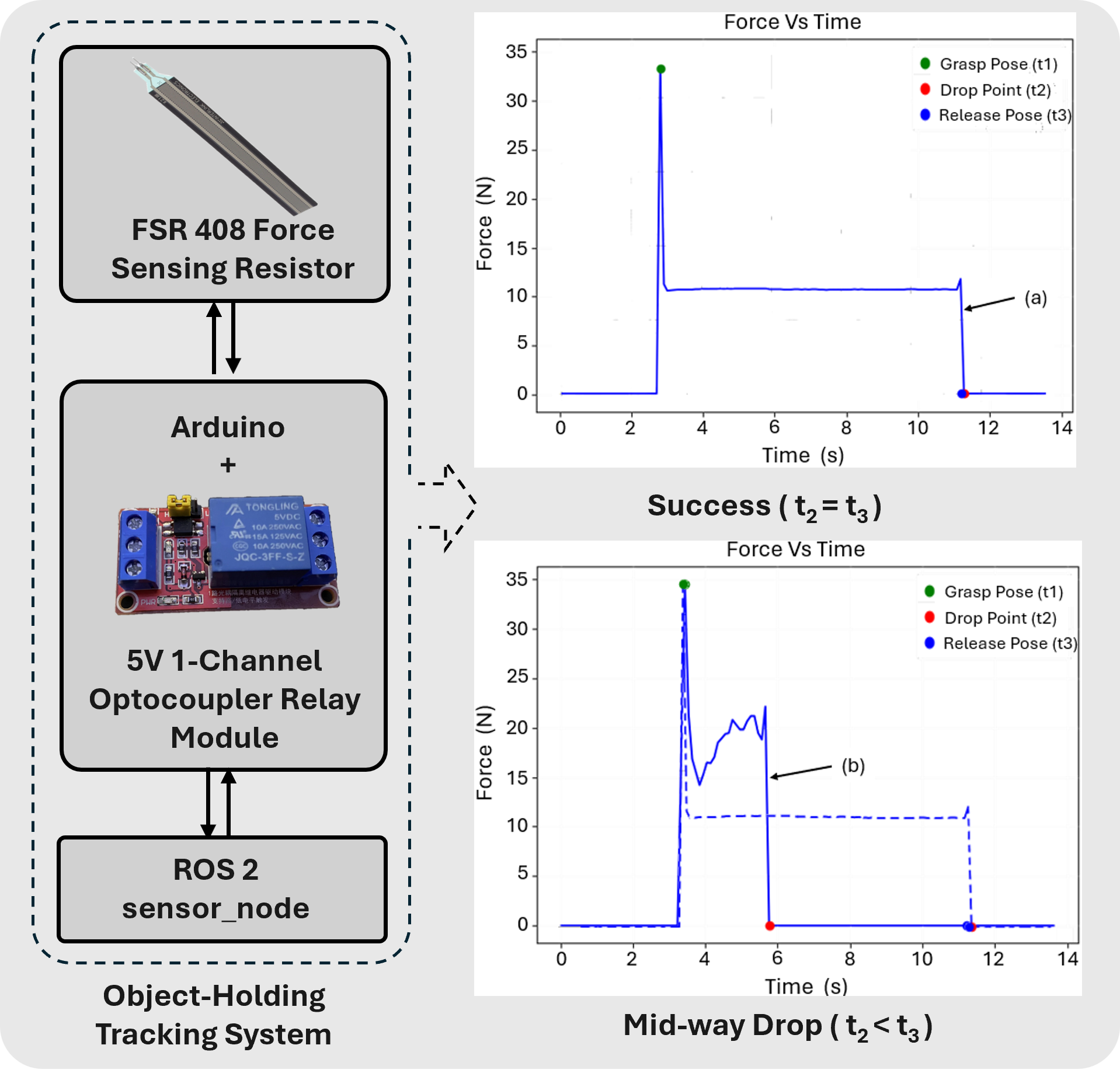}
  \caption{Object-holding tracking system for grasp stability evaluation, showing force sensor measurements during grasp execution: (a) successful transfer, where the object is retained from grasp pose ($t_1$) to release pose ($t_3$) with $t_2 = t_3$, and (b) mid-way drop, where the object is released prematurely due to slippage ($t_2 < t_3$)}
  \vspace{-10pt}
  \label{stability}
\end{figure}

The functionality of the OHT system follows a layered process: once the robot reaches the grasp pose, the relay-control node in the ROS 2 control stack activates the sensor by sending a switching signal through the Arduino. The sensor is deactivated similarly once the robot reaches the release pose. The sensor continuously records force readings from the grasp pose through to the release pose (Figure \ref{graspcycle}). If the object is dropped mid-way due to slippage during transport, the sensor detects a reduction of force to zero at the drop point. The stability score, $S_{b} \in [0,1]$ (refer to Figure \ref{stability}) is defined as the ratio of the time interval from the grasp pose ($t_{1}$) to the drop point ($t_{2}$) over the total expected holding time from the grasp pose ($t_{1}$) to the release pose ($t_{3}$). A value of $S_{b}=1$ indicates that the object is retained throughout the entire execution without slippage.
\vspace{-6pt}
\[
S_{b} = \dfrac{t_{2} - t_{1}}{t_{3} - t_{1}}, \quad \text{where } t_{1} \leq t_{2} \leq t_{3} 
\tag*{Equation (11)}
\]
\vspace{-8pt}

For the suction gripper, an analogous approach is used. The relay circuit controls a Honeywell APB pressure sensor, which records negative pressure during grasp execution. The resulting signal profile mirrors the force-based measurements (inverted due to negative pressure), enabling consistent evaluation of grasp stability across all gripper modalities.

\vspace{-8pt}
\subsubsection{Grasp Efficiency Score ($S_{f}$)}

The efficiency score $S_{f}$ measures how effectively a grasping activity is executed by the gripper. It is determined by the time required to complete the full grasping activity, starting from the pre-grasp position and ending with the release of the object at the release pose (Figure \ref{graspcycle}). To ensure comparability across trials, the grasping time for each trial is normalized using the minimum and maximum grasping times observed across all successful trials within the experiment. This metric captures not only the speed of execution but also reflects the overall responsiveness of the grasping system. 

\begin{equation}
S_{f} = 1 - \frac{T_{c} - T_{\text{min}}}{T_{\text{max}} - T_{\text{min}}}
\tag*{Equation (12)}
\end{equation}

where $T_{c}$ is the grasping activity time for a given trial, $T_{\text{min}}$ is the minimum grasping activity time across successful trials for a experiment, and $T_{\text{max}}$ is the maximum grasping activity time across successful trials for a experiment, with $S_{f} \in [0,1]$.

\vspace{-8pt}
\subsection{Experimental Procedure}\label{sec4}

This section presents the systematic experimental procedure conducted to evaluate grasping performance under varying environmental complexities. As explained in the \ref{BenchPipeline}, the experimental platform consists of a UR10 robotic arm, an Intel RealSense D415 camera, and a test workspace of size $40 \times 40$ cm$^{2}$, which is designed to match the workspace mask of the 6D grasp pose detection models when the camera is positioned 52.5 cm perpendicular to the workspace. We performed four experiments with progressively increasing levels of clutter (L1–L4), as illustrated in Figure~\ref{conceptualization}. These experiments enabled us to highlight critical limitations and demonstrate the necessity of the given grasping modalities. 

\vspace{-8pt}
\subsubsection{Level 1 – Scene with a Single Isolated Inorganic Object}

As a baseline, the first experiment focused on grasping a single isolated object within the scene, where the scene refers to the workspace area captured by the vision system. In this setup, each object from the dataset was randomly positioned in five distinct locations (four corners and center) within the workspace, and grasp attempts were performed. For each object category, this produced 15 trials (3 objects × 5 positions). In total, 105 trials were conducted across all seven categories for a single grasping modality, and 315 trials overall when repeated across all three gripper modalities. An important observation, discussed further in Section~\ref{sec5}, is that the mean success score for the plastic bag and mesh bag categories with the suction gripper was zero. This reflects a systematic limitation of suction-based grasping, where the inability to form a stable seal on highly deformable or porous objects leads to air leakage and poor surface conformity, thereby preventing successful grasps. Consequently, in the subsequent cluttered-scene experiments, these two categories were excluded from the suction gripper evaluation.

\vspace{-8pt}
\subsubsection{Level 2 – Inorganic Objects in Medium-Clutter Scene}

The second experiment increased the clutter level of the scene by introducing multiple objects simultaneously. Using the full dataset (Figure~\ref{dcd}), three clusters were constructed, each consisting of seven objects with representation from all categories. For suction gripper evaluation, a controlled configuration with five objects was used. In each trial, objects within a cluster were randomly dropped from a storage box into the workspace to replicate realistic cluttered conditions. For each cluster, seven consecutive grasp attempts were carried out, selecting the highest-ranked pose from the top eight poses generated by AnyGrasp or SuctionNet after post-processing and grasp-selection. Each cluster was repeated five times, yielding 105 trials for both the rigid and Fin-Ray grippers (7 objects × 5 attempts × 3 clusters) and 75 trials for the suction gripper (5 objects × 5 attempts × 3 clusters). In total, this yields 285 trials across all three grasping modalities. 

\vspace{-8pt}
\subsubsection{Level 3 – Inorganic Objects in Heavy-Clutter Scene}

In the third experiment, we increased the clutterness to high by introducing 14 objects for a scene (This appeared to be the maximum accompanied within the test workspace boundaries and vision-related optimal depth ranges). For these trials, clusters developed in the previous experiment were combined in pairs to form groups of 14 objects, which were then randomly dropped into the workspace to replicate heavy cluttered conditions.  Each group contained two representatives from each category, except for the suction gripper, where two categories were excluded, resulting in 10 objects per scene. For each group, 14 consecutive grasp attempts were carried out, following the same procedure as in Experiment 2. Each group was tested over five repetitions, yielding a total of 210 trials for the rigid and Fin-Ray grippers (14 objects × 5 attempts × 3 groups) and 150 trials for the suction gripper (10 objects × 5 attempts × 3 groups). Across all three gripper modalities, this amounted to 570 trials in total. 

\vspace{-8pt}
\subsubsection{Level 4 – Heavy-Clutter Mixed Food Waste Scene}

The final experiment was designed to replicate realistic food waste sorting conditions by combining seven inorganic objects with seven organic food waste items with a watery base, resulting in a total of 14 objects per scene. This mixed setup introduced additional challenges such as occlusions, adhesion on sticky surfaces, variable textures, including thin watery layers forming between the gripper surface and the object and entanglement effects, all of which reflect industrial conditions within a controlled laboratory environment. Following the same procedure as in level 3, objects were randomly dropped into the workspace, and 14 consecutive grasping cycles were performed, each targeting a single object, with the exception applied to the suction gripper as noted previously. The complete grasping pipeline was executed across all three gripper modalities, yielding a total of 570 trials for evaluating performance in these heterogeneous scenes. In all four experiments, for every grasp attempt, regardless of success or failure all six evaluation metrics, including the graspability parameters and grasp performance metrics, were systematically recorded. 

\section{Results and Discussion}\label{sec5}

In this section, the results analysis is presented in three phases. First, we compare the grasp performance of the three unimodal grippers across four high-fidelity experiments, aiming to identify category-wise excellence. Second, we analyze how pre-grasp conditions expressed through graspability parameters influence performance measures for each gripper modality. Finally, we conduct a grasp failure analysis in two stages: we first examine the occurrence rates of failure modes, and then assess the impact of graspability parameters on the identified failures to inform future gripper improvements. 

\vspace{-8pt}
\subsection{Performance Comparison of Grasping Modalities}\label{sec5_1}

Based on the performance scores for the three metrics, we use a radar chart to visualize and compare the overall performances of the different grippers, with the primary goal of identifying object category-wise performance across the four levels of experiments (Table ~\ref{radar}). The radar chart has three axes, representing  grasp success $S_{p}$, grasp stability $S_{b}$, and grasp efficiency $S_{f} \in [0,1]$. For each experimental level, category-wise mean scores were computed and plotted on the corresponding axes. Connecting these points forms a polygon that illustrates the overall performance profile of each gripper. A vertex positioned at the centre of the chart indicates a score of zero on that axis, whereas outward extensions represent higher scores. In some cases, the axes are contracted rather than starting strictly at zero, to better highlight relative differences between grippers. 

\begin{table*}[!htbp]
\centering
\vspace{-6pt}
\caption{Evaluation of grasp performance using radar charts. Performance polygons are color-coded by gripper (blue: rigid, pink: Fin-Ray, green: suction), where the gripper with the largest enclosed area represents the highest overall performance, indicated by the corresponding background color.}
\label{radar}
\vspace{-6pt}
\small
\setlength{\tabcolsep}{2pt}
\renewcommand{\arraystretch}{0.95}

\begin{tikzpicture}
\node[inner sep=0pt] (T) {
\begin{tabular*}{0.95\textwidth}{@{\extracolsep\fill}
  Y{0.09\textwidth} Y{0.2\textwidth} Y{0.2\textwidth} Y{0.2\textwidth} Y{0.2\textwidth}
  @{\extracolsep\fill}}
\toprule
\textbf{Object} & \multicolumn{4}{c}{\textbf{Experiment Level}} \\
\cmidrule{2-5}
\textbf{category} & \textbf{Level 1} & \textbf{Level 2} & \textbf{Level 3} & \textbf{Level 4} \\
\midrule

Plastic Bag &
\includegraphics[width=0.190\textwidth]{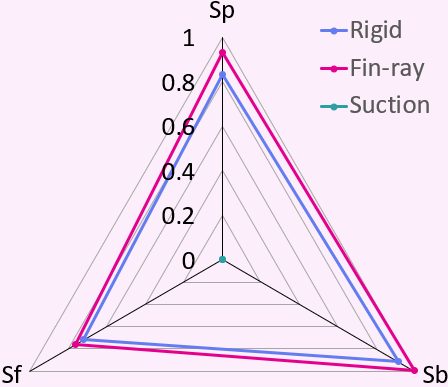} &
\includegraphics[width=0.190\textwidth]{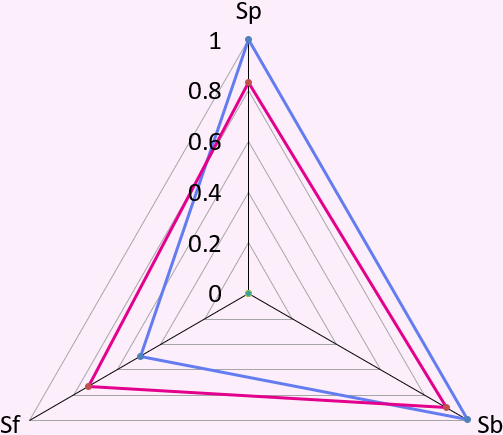} &
\includegraphics[width=0.190\textwidth]{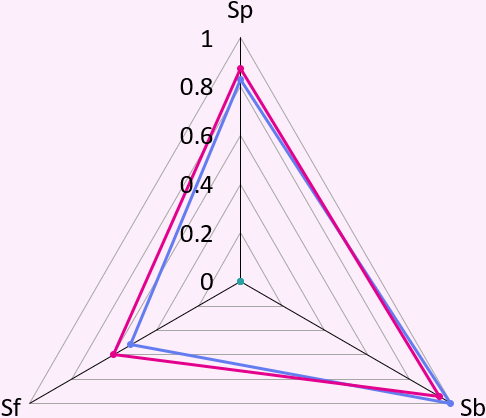} &
\includegraphics[width=0.190\textwidth]{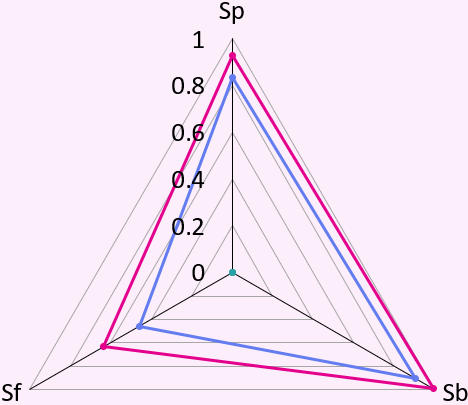} \\

Mesh Bag &
\includegraphics[width=0.190\textwidth]{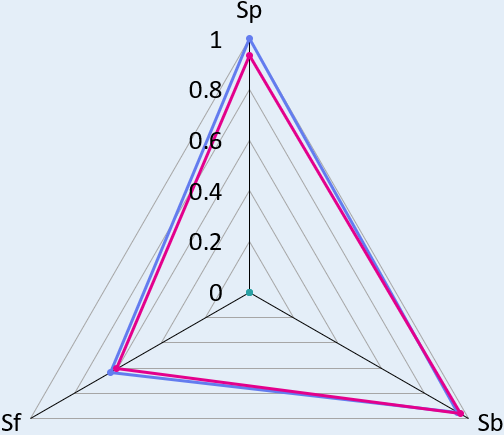} &
\includegraphics[width=0.190\textwidth]{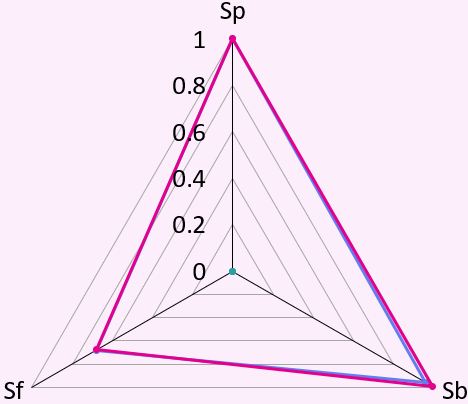} &
\includegraphics[width=0.190\textwidth]{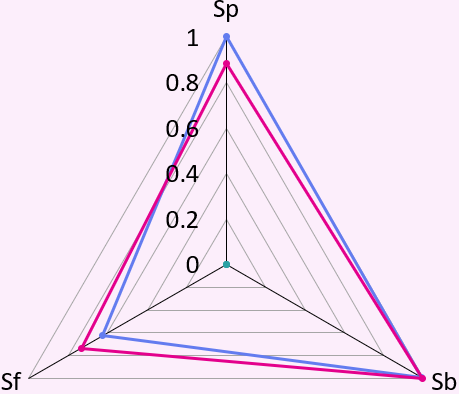} &
\includegraphics[width=0.190\textwidth]{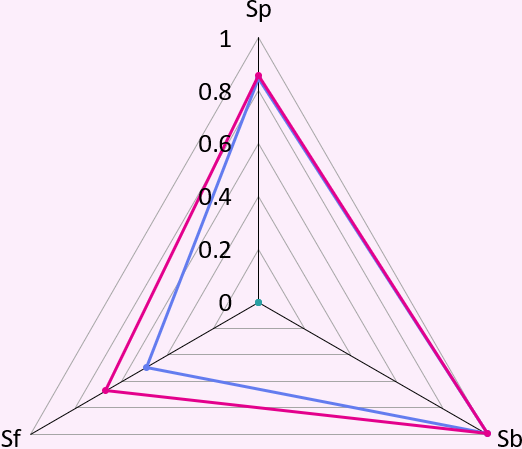} \\

LPB &
\includegraphics[width=0.190\textwidth]{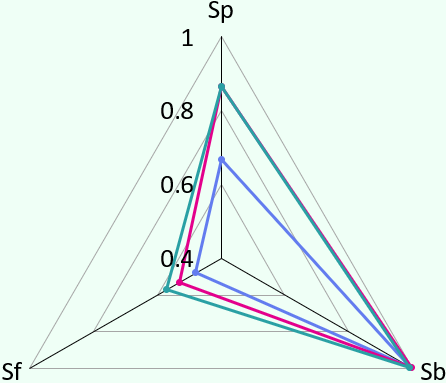} &
\includegraphics[width=0.190\textwidth]{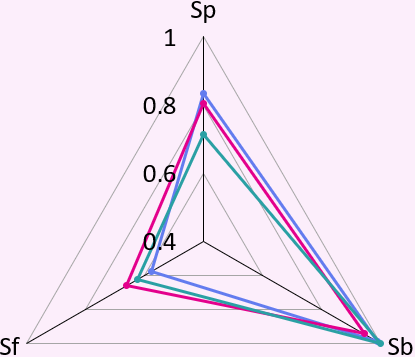} &
\includegraphics[width=0.190\textwidth]{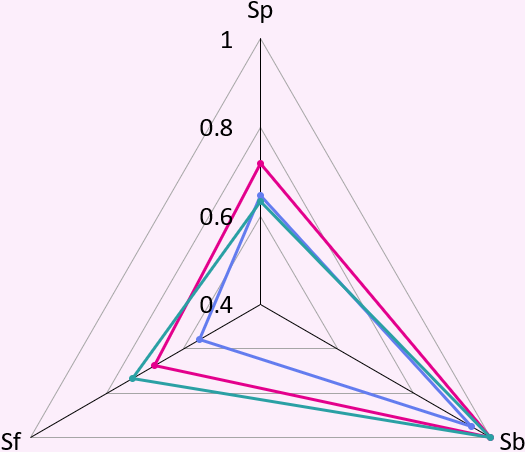} &
\includegraphics[width=0.190\textwidth]{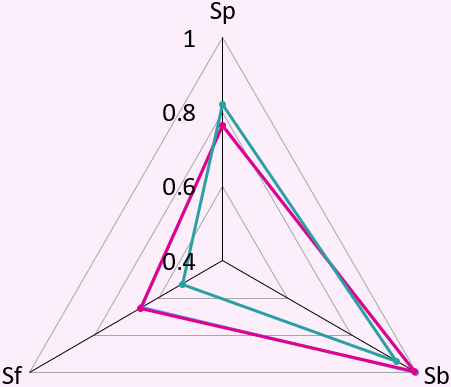} \\

Plastic Container &
\includegraphics[width=0.190\textwidth]{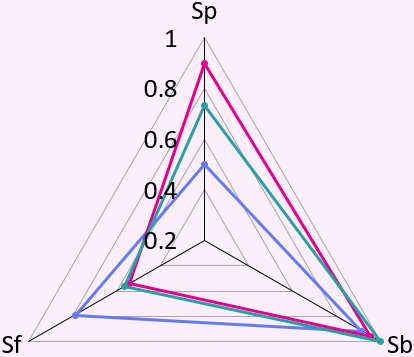} &
\includegraphics[width=0.190\textwidth]{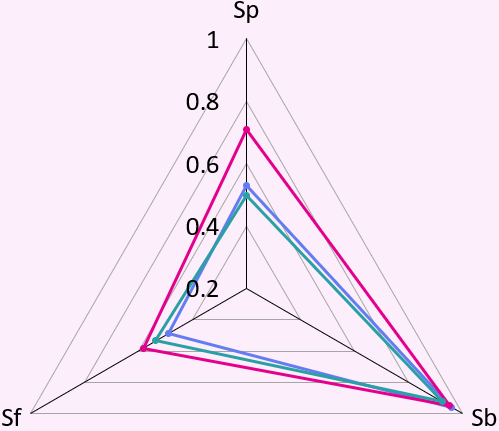} &
\includegraphics[width=0.190\textwidth]{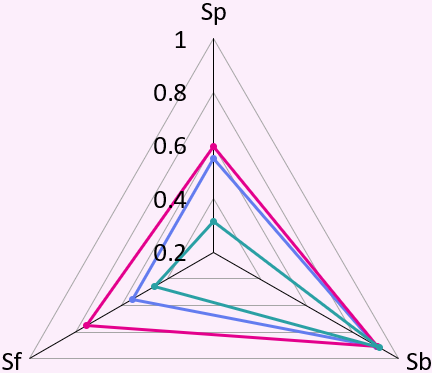} &
\includegraphics[width=0.190\textwidth]{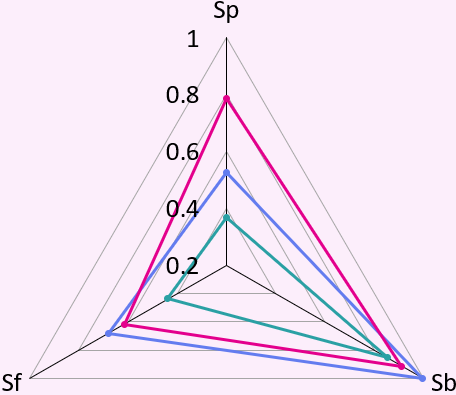} \\

Plastic Plate &
\includegraphics[width=0.190\textwidth]{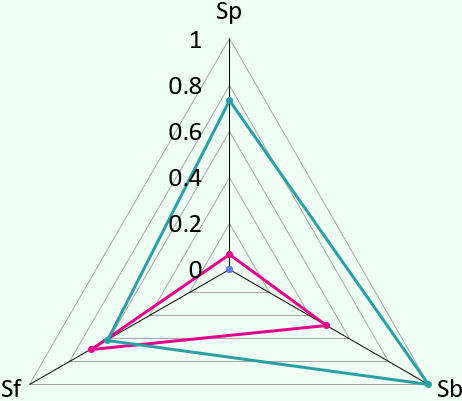} &
\includegraphics[width=0.190\textwidth]{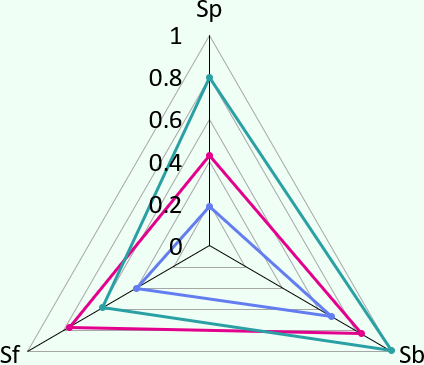} &
\includegraphics[width=0.190\textwidth]{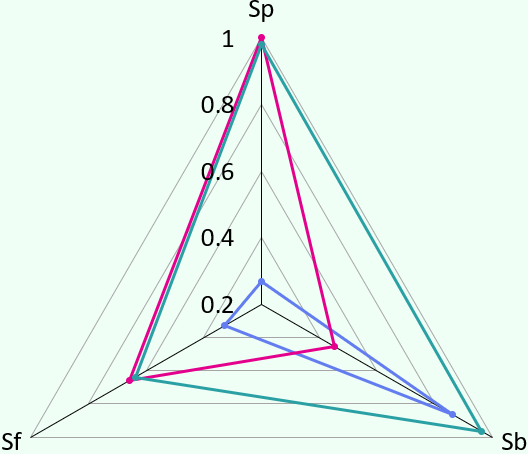} &
\includegraphics[width=0.190\textwidth]{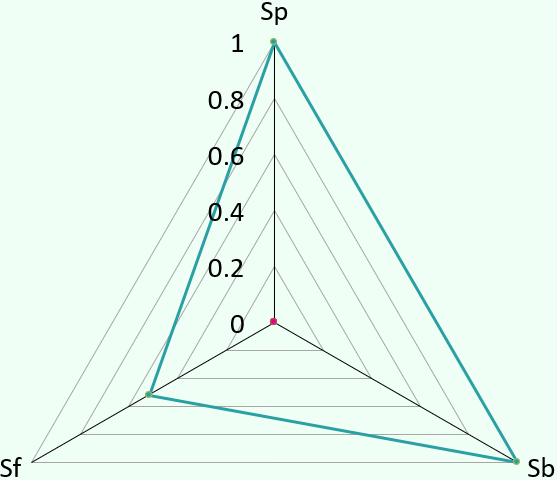} \\

Plastic Bottle &
\includegraphics[width=0.190\textwidth]{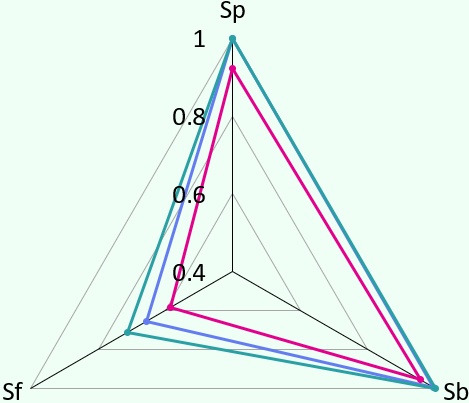} &
\includegraphics[width=0.190\textwidth]{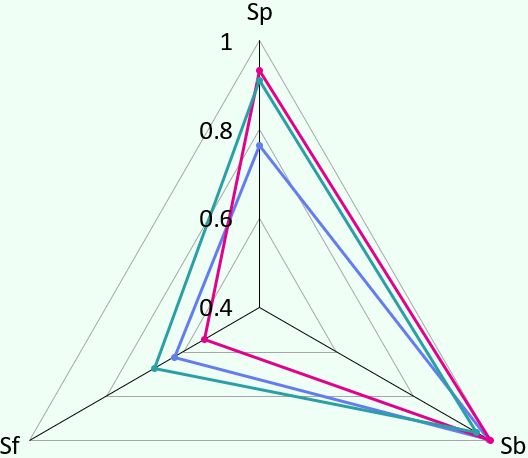} &
\includegraphics[width=0.190\textwidth]{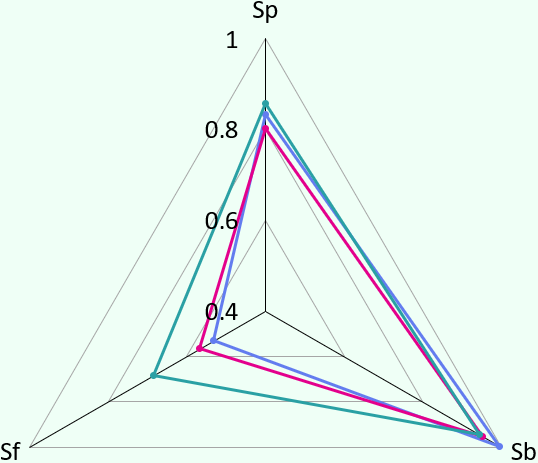} &
\includegraphics[width=0.190\textwidth]{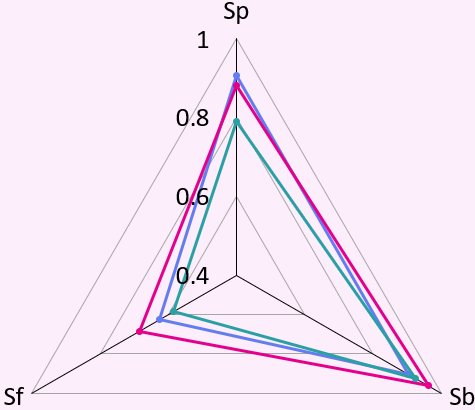} \\

Tin Can &
\includegraphics[width=0.190\textwidth]{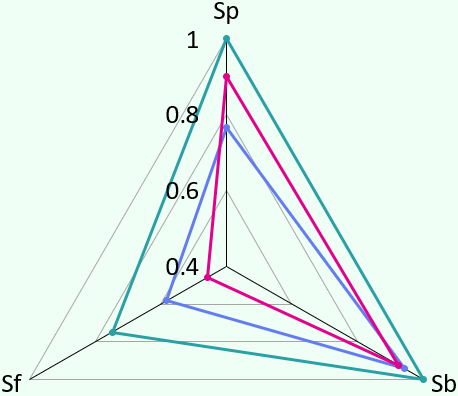} &
\includegraphics[width=0.190\textwidth]{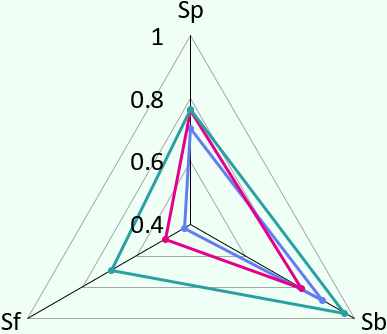} &
\includegraphics[width=0.190\textwidth]{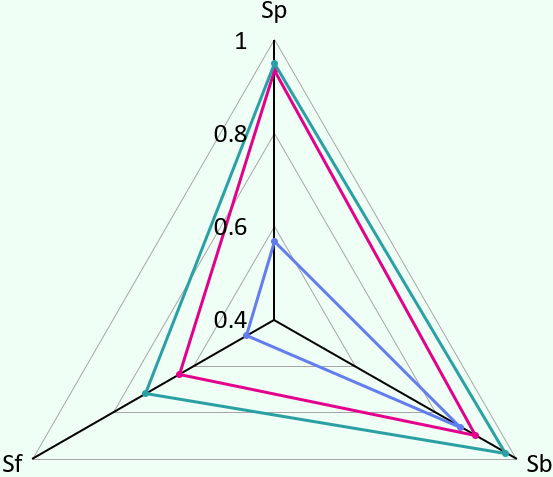} &
\includegraphics[width=0.190\textwidth]{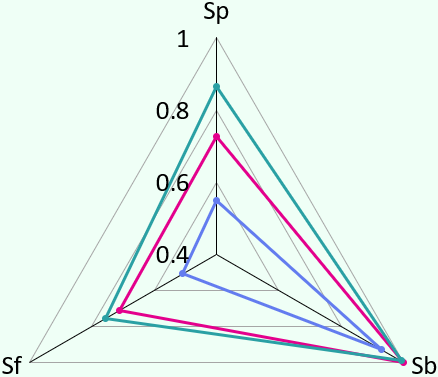} \\

\bottomrule
\end{tabular*}
};

\draw[very thick,<->,color=red!70]
([xshift=-0.2cm,yshift=10cm]T.west) --
([xshift=-0.2cm,yshift=-11cm]T.west);

\node[rotate=90,font=\small] at ([xshift=-0.5cm,yshift=10cm]T.west) {Deformable};
\node[rotate=90,font=\small] at ([xshift=-0.5cm,yshift=-11cm]T.west) {Rigid};

\end{tikzpicture}
\end{table*}

\begin{description}
\item[Level 1] Analysis of the radar chart polygons revealed that the Fin-Ray gripper achieved the best performance with plastic bags and plastic containers, while the rigid gripper excelled only in mesh bags. In contrast, the suction gripper showed the lowest performance in these deformable categories, with plastic bags and mesh bags yielding zero success due to non-conforming surfaces that prevented suction sealing. A noteworthy observation is that in single isolated object scenarios, the suction gripper outperformed the other grippers in four out of seven categories (laminated paper boxes (LPB), plastic plates, plastic bottles, and tin cans), highlighting its suitability for flat, rigid surfaces. In these categories, the other two grippers showed reduced success, and the rigid gripper in particular recorded zero success for plastic plates due to collisions with the base when the object was positioned too close to the surface (further explained in \ref{sec5_3}).
\vspace{4pt}

\item[Level 2 and 3] Both the medium and high clutter experiments (7 objects and 14 objects per scene, respectively) demonstrated similar overall performance despite the increase in scene complexity. Among the seven object categories, the Fin-Ray gripper outperformed in four categories: plastic bags, mesh bags, laminated paper boxes (LPB), and plastic containers. The corresponding radar charts indicate that while both the rigid gripper and the Fin-Ray gripper performed similarly in these categories, the Fin-Ray consistently outperformed the rigid gripper due to its excellent shape-morphing capability, which adapts effectively to varying object surfaces. However, both grippers exhibited very low overall performance on flat and rigid surfaces. Consistent with the results from the isolated scenario, the suction gripper achieved the highest overall performance in plastic plates, plastic bottles, and tin cans, the three categories characterized by high rigidity and flat surface structures.
\vspace{4pt}

\item[Level 4] In the mixed food waste scenario, the Fin-Ray gripper achieved the highest overall success, outperforming the other two grippers and demonstrating greater tolerance to the additional challenges introduced by organic waste. This superior performance was observed across five of the seven categories: plastic bags, mesh bags, laminated paper boxes (LPB), plastic containers, and plastic bottles.  A notable observation is that while the suction gripper outperformed in plastic bottles during inorganic cluttered scenes (Experiments 2 and 3), the advantage shifted to the Fin-Ray gripper in the mixed waste scene, highlighting the suction gripper’s inability to tolerate varying object geometries and fluidic conditions that disrupt suction. Nevertheless, consistent with other cluttered experiments, the suction gripper still excelled in plastic plates and tin cans, benefiting from its strong conformance on smooth, thin flat surfaces, despite disturbances caused by thin liquid films between the suction cup and object surfaces.
\end{description}\medskip

Overall, despite the challenges posed by cluttered scenes, the adaptive Fin-Ray gripper demonstrated the highest resilience when handling deformable to semi-deformable objects, such as plastic bags, mesh bags, laminated paper boxes, plastic containers, and plastic bottles, indicating strong cross-category suitability for the given use case. In particular, the radar chart results for L4 show that it outperforms the other two grippers when dealing with the sticky and irregular nature of food-waste scenarios. However, the Fin-Ray gripper showed lower resilience with objects that had thin, flat surfaces (plastic plates and tin cans), where frequent collisions with the work base became a significant issue due to the stiffness of its fingertips. In contrast, for these object categories, the suction gripper exhibited the highest adaptability, owing to its strong conformance with flat surfaces and its top-down approach. Unlike the rigid and Fin-Ray grippers, which operate in parallel to the work surface, the suction gripper’s vertical approach minimizes base collisions, making it particularly effective for thin, flat objects.

\vspace{-4pt}
\subsection{Effect of Graspability on Grasp Performance} \label{sec5_2}

\begin{figure*}[!h]
  \centering
  \vspace{-14pt}
  \includegraphics[width=0.7\textwidth,height=0.31\textheight,keepaspectratio]{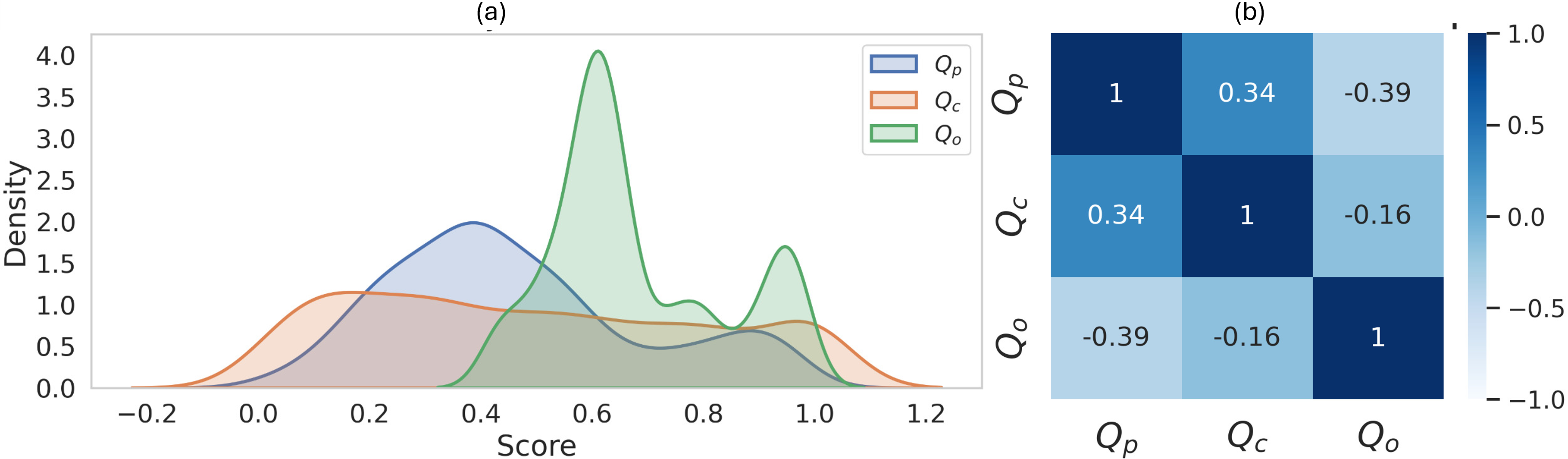}
  \caption{Exploratory data analysis of graspability parameters. (a) Kernel density distributions of object score $Q_{o}$, pose score $Q_{p}$ and clutter score $Q_{c}$ (b) Correlation heatmap}
  \vspace{-10pt}
  \label{EDA}
\end{figure*}

In this phase, we analyse how pre-grasp conditions (graspability parameters), identified as the primary predictors of grasp performance, influence performance in each gripper modality during grasping in cluttered GIC scenarios (L2–L4). We use inferential modeling to explain and quantify how the predictors, pose score $Q_{p}$, object score $Q_{o}$, and clutter score $Q_{c}$ impact the three independent performance outcomes: grasp success $S_{p}$, grasp stability $S_{b}$, and grasp efficiency $S_{f}$. As the first step, we conducted Exploratory Data Analysis (EDA) to understand the suitability of the models for the obtained results. We examined the distributions of the graspability parameters using histograms and correlation density plots. The predictors show non-normal but only mildly skewed distributions and are already represented on an interpretable [0,1] scale. Correlations between predictors are low ($|r| < 0.4$), indicating no multicollinearity and therefore, logistic-based models are preferred over linear models (Figure ~\ref{EDA}). 
\vspace{4pt}

In our dataset, $S_{p}$ serves as a binary indicator, while $S_{b}$ and $S_{f}$ are continuous rates constrained within $[0,1]$. Accordingly, each outcome is modeled using a logit link appropriate to its scale: $S_{p}$ is fit with a standard logistic regression \cite{hosmer2013} whereas $S_{b}$ and $S_{f}$ are modeled using a fractional logit regression that explicitly accounts for the fractional nature of the dependent variable \cite{Swathi2022}. The same set of predictors is used across all three models: $Q_o$, $Q_p$, and $Q_c$ with gripper type included as a categorical factor, $\text{C}(\text{Gripper\_type})$, where Rigid is set as the reference (baseline) category. Interactions between each graspability parameter and gripper type are also incorporated. Formally, for each outcome $S_k \in \{S_p,S_b,S_f\}$, the index $k$ denotes the specific performance metric being modeled (success, stability, or efficiency). The model is expressed as:
\vspace{-14pt}
\begin{equation}
\begin{split}
g_k\!\left(\mathbb{E}[S_k]\right) 
&= \beta_{0,k} + \beta_{o,k} Q_o  + \beta_{p,k} Q_p +  \beta_{c,k} Q_c + \text{C}(\text{Gripper\_type}) \\
&+ (Q_o+Q_p+Q_c)\!:\!\text{C}(\text{Gripper\_type})
\end{split}
\tag*{Equation (13)}
\end{equation}
\vspace{-12pt}

The function $g_k(\cdot)$  denotes the appropriate link function applied to the expected value of each outcome. The coefficients $\beta_{o,k}$, $\beta_{p,k}$, and $\beta_{c,k}$ represent the effect of the graspability parameters $Q_o$, $Q_p$, and $Q_c$ on the $k$-th outcome, respectively. The terms $\beta_{o,k} Q_o$, $\beta_{p,k} Q_p$, and $\beta_{c,k} Q_c$ therefore capture the contribution of object quality, pose quality, and clutter level to the predicted performance outcome. The interaction terms $(Q_o + Q_p + Q_c):\text{C}(\text{Gripper\_type})$ are essential, as they allow these effects to vary across gripper types relative to the baseline (Rigid). For example, the benefit of improved pose quality may be greater for suction than for rigid (the baseline), rather than being restricted to a common slope across all grippers. We deliberately exclude the experimental level factor from the mean model. Level is a coarse design factor that largely proxies aspects already measured continuously by the graspability parameters $Q_o$, $Q_p$, and $Q_c$. For all three outcomes, we report regression coefficients, their exponentiated form (odds ratios), p-values, and 95\% confidence intervals. For binary success $S_{p}$, the odds ratios have the usual interpretation in terms of odds of success. For stability $S_{b}$ and efficiency $S_{f}$, which are fractional outcomes, the odds ratios describe multiplicative effects on the odds of the expected proportion. 
\vspace{4pt}

Considering the odds ratios and p-values (Table \ref{regression}) from the regression models, in overall, the Fin-ray (refer to row $Finray$ vs $Rigid$) and Suction (refer to row $Suction$ vs $Rigid$) grippers show significantly improved performance relative to the Rigid baseline ($p<0.05$). For grasp success $S_{p}$, object quality $Q_o$ emerges as the dominant driver for the rigid (refer to row $Q_o$)  and suction gripper (refer to row $Q_o$ x $Suction$). For rigid higher $Q_o$  significantly increases the odds of a successful grasp  ($p=0.0008<0.05$). In contrast, for the Fin-ray gripper, the effect of $Q_o$ is weak, whereas for the Suction gripper, it shows a strong negative association. This trend is consistent across stability $S_{b}$ and efficiency $S_{f}$, where $Q_o$ continues to be the most influential predictor.  
\vspace{4pt}

\definecolor{lg}{HTML}{D9F2D9} 
\definecolor{lr}{HTML}{F8D6D6} 

\begin{table*}[!h]
\centering
\vspace{-10pt}
\scriptsize
\setlength{\tabcolsep}{3.5pt}
\caption{Regression results for grasp performance metrics. Grasp success ($S_{p}$) is modeled using logistic regression, while stability ($S_{b}$) and efficiency ($S_{f}$) use fractional logit models. The Rigid gripper is the reference category. Green = significant positive ($p<0.05$), Red = significant negative ($p<0.05$), White = not significant.}
\vspace{-8pt}
\begin{tabular}{lccc ccc ccc}
\toprule
 & \multicolumn{3}{c}{$S_{p}$} & \multicolumn{3}{c}{$S_{b}$} & \multicolumn{3}{c}{$S_{f}$ } \\
\cmidrule(lr){2-4}\cmidrule(lr){5-7}\cmidrule(lr){8-10}
\textbf{Term} & coef & OR & p & coef & OR & p & coef & OR & p \\
\midrule
$Q_p$ &
1.4543 & 4.2815 & 0.1656 &
1.3233 & 3.7557 & 0.1909 &
0.3579 & 1.4304 & 0.6345 \\

$Q_c$ &
1.1525 & 3.1661 & 0.0790 &
1.1121 & 3.0408 & 0.0897 &
0.6832 & 1.9802 & 0.0400 \\

$Q_o$ &
\cellcolor{lg} 4.0842 & \cellcolor{lg} 59.3952 & \cellcolor{lg} 0.0008 &
\cellcolor{lg} 4.0813 & \cellcolor{lg} 59.2247 & \cellcolor{lg} 0.0004 &
\cellcolor{lg} 2.0424 & \cellcolor{lg} 7.7091 & \cellcolor{lg} 0.0101 \\

\addlinespace[2pt]
Finray vs Rigid &
\cellcolor{lg} 1.6127 & \cellcolor{lg} 5.0165 & \cellcolor{lg} 0.0341 &
\cellcolor{lg} 1.4474 & \cellcolor{lg} 4.2522 & \cellcolor{lg} 0.0331 &
\cellcolor{lg} 1.1662 & \cellcolor{lg} 3.2098 & \cellcolor{lg} 0.0293 \\

$Q_p \times$ Finray &
-0.4471 & 0.6395 & 0.6879 &
-0.3319 & 0.7175 & 0.7465 &
0.1092 & 1.1154 & 0.8907 \\

$Q_c \times$ Finray &
0.0252 & 1.0255 & 0.9765 &
0.1091 & 1.1153 & 0.8948 &
-0.4024 & 0.6687 & 0.3189 \\

$Q_o \times$ Finray &
-1.5309 & 0.2163 & 0.1231 &
-1.4326 & 0.2387 & 0.1291 &
-0.9992 & 0.3682 & 0.1661 \\

\addlinespace[2pt]
Suction vs Rigid &
\cellcolor{lg} 5.6499 & \cellcolor{lg} 284.2757 & \cellcolor{lg} 0.0415 &
\cellcolor{lg} 6.4123 & \cellcolor{lg} 609.2784 & \cellcolor{lg} 0.0062 &
\cellcolor{lg} 3.1153 & \cellcolor{lg} 22.5391 & \cellcolor{lg} 0.0337 \\

$Q_p \times$ Suction &
-0.2512 & 0.7779 & 0.8420 &
-0.2453 & 0.7825 & 0.8408 &
0.9618 & 2.6163 & 0.2646 \\

$Q_c \times$ Suction &
-1.7267 & 0.1779 & 0.1530 &
-1.9121 & 0.1478 & 0.1294 &
-1.1258 & 0.3244 & 0.0666 \\

$Q_o \times$ Suction &
\cellcolor{lr} -7.5821 & \cellcolor{lr} 0.0005 & \cellcolor{lr} 0.0431 &
\cellcolor{lr} -8.4816 & \cellcolor{lr} 0.0002 & \cellcolor{lr} 0.0081 &
\cellcolor{lr} -4.7211 & \cellcolor{lr} 0.0089 & \cellcolor{lr} 0.0124 \\

\vspace{-16pt}
\end{tabular}
\label{regression}
\end{table*}

The pose quality score $Q_p$ and clutter metric $Q_c$ exhibit comparatively moderate effects across all three outcomes. This indicates that pose quality does not strongly differentiate performance, as grasp candidates are pre-filtered through grasp detection and collision checking prior to execution, thereby reducing variability in $Q_p$. Similarly, the influence of $Q_c$ appears limited, as it captures global scene density but does not explicitly encode higher-order spatial relationships. In addition, the adaptive characteristics of the Fin-ray gripper and the localized contact mechanism of suction reduce sensitivity to clutter, enabling stable performance even in densely populated scenes.
\vspace{4pt}

\begin{figure*}[!t]
    \centering
    \vspace{-10pt}
    \includegraphics[width=\textwidth]{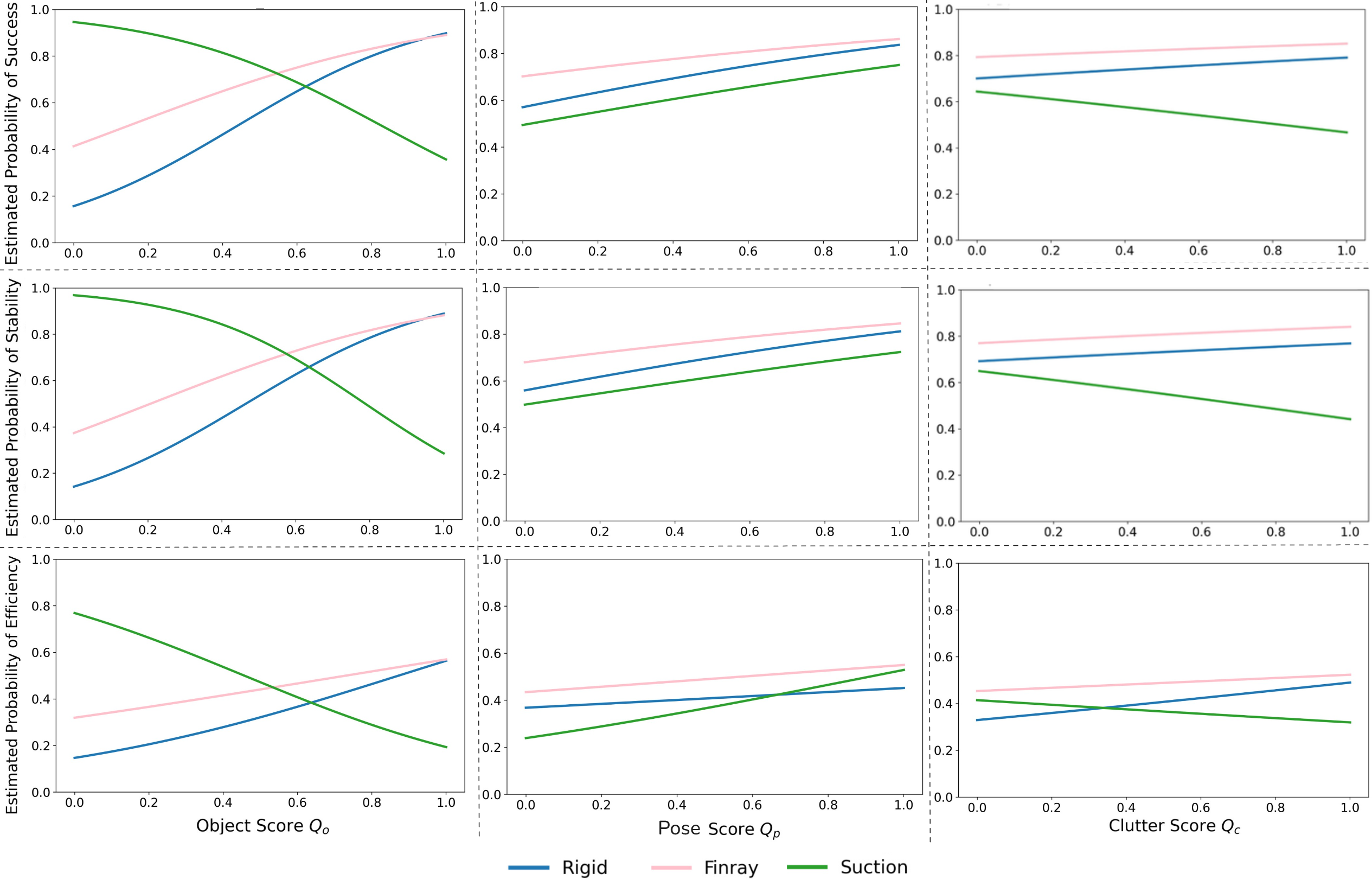} 
    \captionsetup{skip=2pt}
    \caption{Partial dependence plots illustrating the effect of graspability parameters (Object Score $Q_o$, Grasp Score $Q_p$, and Clutter Score $Q_c$) on the predicted probabilities of Grasp Success $S_p$, Grasp Stability $S_b$ and Grasp Efficiency $S_f$ across gripper modalities (Rigid, Fin-ray, and Suction).}
    \vspace{-14pt}
    \label{pdpplots}
\end{figure*}

To further interpret these effects, partial dependence plots (PDPs) (Figure~\ref{pdpplots}) were generated for each performance outcome against each graspability parameter. PDPs are used to visualize the marginal effect of a single predictor on the expected model output while averaging over the distribution of the remaining variables \cite{friedman2001,molnar2020}. In the context of logistic and fractional logit models, PDPs illustrate how changes in a given graspability parameter influence the predicted probability (for $S_{p}$) or expected proportion (for $S_{b}$ and $S_{f}$) after applying the inverse logit transformation. This enables an interpretable representation of model behavior beyond coefficient-level analysis, particularly in the presence of interaction terms between predictors and gripper types.  
\vspace{4pt}

The object score ($Q_o$), representing object-quality related pre-grasp conditions, shows a strong positive influence on grasp success, stability, and efficiency for both the Rigid and Fin-ray grippers, with performance increasing consistently as $Q_o$ rises. In contrast, the Suction gripper exhibits a clear negative relationship across all outcomes, indicating reduced performance for higher $Q_o$ values, which correspond to more deformable/porous objects. For pose quality ($Q_p$), all three grippers demonstrate a moderate positive relationship with performance metrics. The rigid and Fin-ray grippers show a consistent improvement in success, stability, and efficiency as $Q_p$ increases, while the suction gripper exhibits a milder positive trend. This suggests that improved 6D grasp pose estimation contributes to performance across all modalities, although its effect is less pronounced compared to object quality. For clutter ($Q_c$), the trends vary across grippers. The rigid and Fin-ray grippers show weak or near-negligible relationships with performance, indicating robustness to global scene density. In contrast, the suction gripper demonstrates a consistent negative trend across all outcomes, highlighting its sensitivity to clutter and reduced effectiveness in densely populated scenes.
\vspace{4pt}

Overall, these observations do not diminish the relevance of $Q_p$ and $Q_c$, but rather highlight that object quality $Q_o$, which directly governs physical interaction dynamics, remains the primary determinant of grasp success. Furthermore, these findings reveal a hierarchy of influence among graspability parameters under realistic operating conditions. While $Q_o$ governs the fundamental feasibility of physical interaction, $Q_p$ and $Q_c$ act as a secondary factor whose effects are compensated by advanced 6D grasp vision. This distinction is critical: unlike existing methods that provide a unified difficulty score, GRAB offers a decomposed and physically interpretable representation of graspability, enabling the identification of dominant factors within application-specific grasping process.

\vspace{-6pt}
\subsection{Failure Mode Analysis}\label{sec5_3}

This section presents the third phase of the results analysis, conducted in two stages. The first stage focuses on identifying the most frequent failure modes across the three gripper modalities, while the second stage examines how graspability parameters (pre-grasp conditions) contribute to the occurrence of these failures.

\vspace{-6pt}
\subsubsection{Frequency of Grasp Failure Modes}\label{sec5_3_1}

We categorized failure modes in to three major categories: physical interaction failures, perception failures, and execution failures. Physical interaction failures included gripper collisions with the work base or targeted object (CL), collisions with surrounding objects (CLS), slippage (SL), and failures related to object non-conformance due to non-planar or highly irregular surfaces (ON). Perception failures primarily involved grasp pose misalignment due to incorrect pose generation by the vision system (WGP). Execution failures referred to errors arising during the grasp control process that prevented the successful completion of the grasp cycle.  
\vspace{4pt}

\begin{figure*}[!t]
  \centering
  \vspace{-10pt}
  \includegraphics[width=0.9\textwidth,height=0.30\textheight,keepaspectratio]{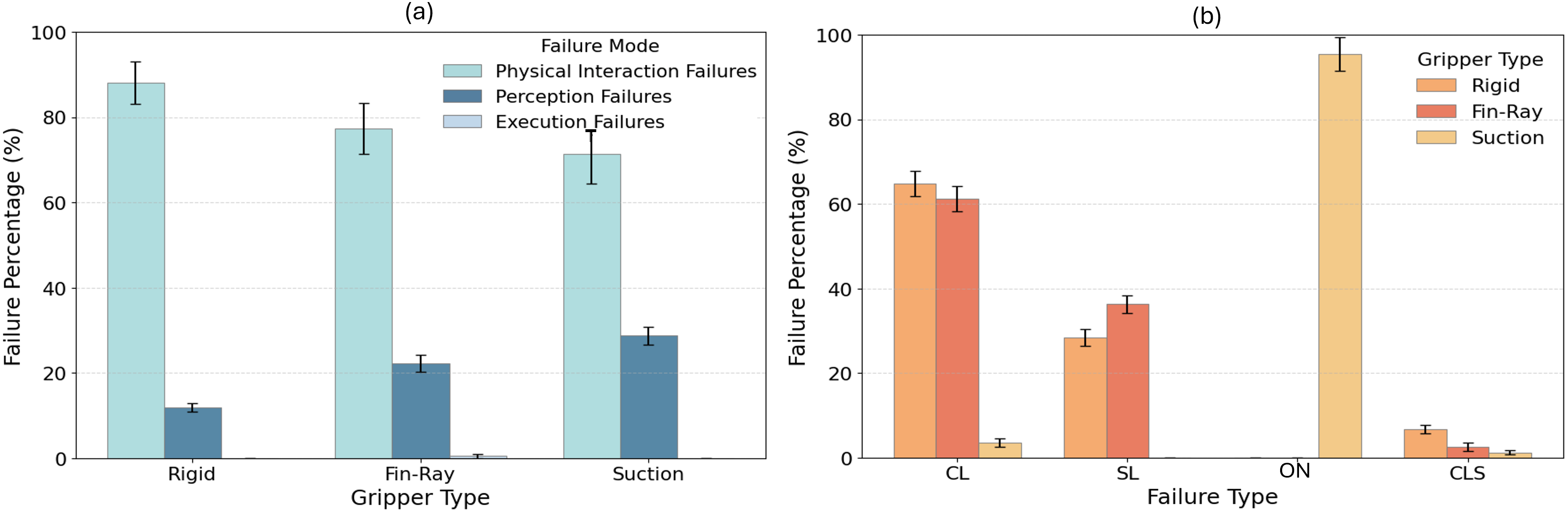}
  \vspace{-8pt}
  \caption{(a) Variation in three major categories of failure modes: physical interaction failures, perception failures, and execution failures observed across the three grasping modalities. (b) Variation of physical interaction failure modes: CL - gripper collisions with the work base or target object, CLS - collisions with surrounding objects, SL - slippage, and ON - failures related to object non-conformance due to non-planar or highly irregular surfaces}
  \vspace{-12pt}
  \label{fmodes}
\end{figure*}

As illustrated  in Figure~\ref{fmodes} (a), physical interaction failures accounted for more than 70\% of the total failure cases across all three grippers, highlighting that improvements in the physical design of grippers are essential. Perception failures represented 11\%, 22\%, and 28\% of the failures for the rigid, Fin-Ray, and suction grippers, respectively, indicating that the 6D grasp pose detection methods employed in the system contributed reasonably well to grasp evaluation by limiting vision-related errors. Execution failures constituted less than 1\% of the total cases and occurred only with the Fin-Ray gripper, which highlights the high accuracy and reliability of the ROS-based path planning algorithm integrated into the benchmarking pipeline.
\vspace{4pt}

As physical interaction failures accounted for more than 70\% of the total, these categories were analyzed in greater detail (Figure~\ref{fmodes}(b)). For the rigid and Fin-Ray grippers, collisions with the work base or the targeted object (CL) were the most frequent, occurring in over 60\% of cases. These failures typically involved the fingertip colliding with either the base surface or the object itself, highlighting the need to improve longitudinal compliance of fingertip to reduce such cases. In contrast, for the Suction gripper, object non-conformance failures (ON) constitute the largest proportion of physical interaction failures. These were mainly caused by the inability of the suction cup to conform to varying object surfaces, particularly under fluidic conditions that closely resemble real food waste sorting environments. This indicates the importance of redesigning the suction cup to better accommodate adversarial object properties and environmental conditions, as ON accounted for nearly 95\% of suction failures. 
\vspace{4pt}

Additionally, slipping (SL) was the second most significant physical interaction failure for both the rigid and Fin-Ray grippers, contributing to more than 25\% of total failures in each case. This was primarily observed in the L4 experiments, where the liquid nature of the environment reduced the coefficient of friction and increased the likelihood of slip within the limited contact regions. The high occurrence of slippage highlights the need to improve the frictional properties of the gripper contact surfaces. In contrast, collisions with surrounding objects (CLS) were relatively rare, accounting for less than 10\% of failures across all gripper modalities. This indicates that the grasp pose generation provided by the selected vision models is sufficiently accurate in avoiding unintended interactions with nearby objects.

\begin{figure*}[!t]
    \centering
    \vspace{-10pt}
    \includegraphics[width=\textwidth]{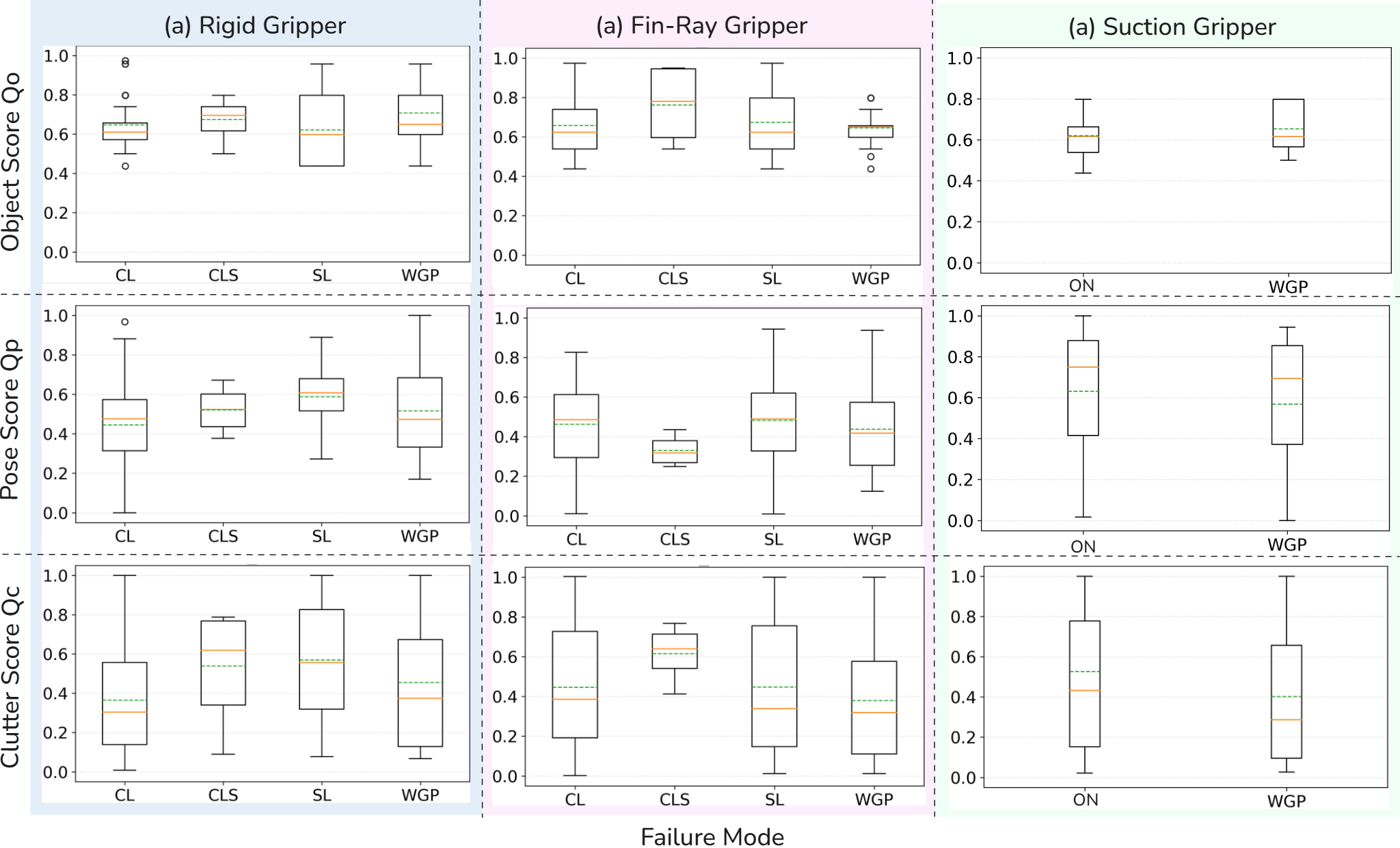} 
    \captionsetup{skip=2pt}
    \caption{Impact of graspability parameters (pre-grasp conditions) $Q_o$, $Q_p$ and $Q_c$ on the occurrence of grasp failures for three gripper modalities: (a) Rigid two-finger parallel gripper (RG2) (b) Soft adaptive two-finger Fin-Ray gripper (c) Suction gripper.}
    \vspace{-14pt}
    \label{boxplots}
\end{figure*}

\vspace{-6pt}
\subsubsection{Impact of Graspability Parameters on Grasp Failure Occurrence}\label{sec5_3_2}

The second stage of the failure analysis involves identifying the impact of pre-grasp conditions on the observed failure modes. In this context, the object score $Q_{o}$, pose score $Q_{p}$, and clutter score $Q_{c}$ are treated as predictors for the failure modes CL, CLS, SL, ON, and WGP, enabling an understanding of which pre-grasp condition contributes most to each mode of failure.  The box plots in Figure~\ref{boxplots} compare the influence of these pre-grasp measures on different failure modes across grippers, and the overall effects are summarized in Table~\ref{failureinfluence}. 
\vspace{4pt}

For rigid and Fin-Ray grippers, collisions with the work base or the target object (CL) were common when the object score $Q_o$ was low, as with rigid items such as plastic containers, plates, or bottles, but not with deformable objects like plastic bags. This strengthens the argument made in Section~\ref{sec5_3_1}, highlighting the need for longitudinal compliance of fingertip. CL is associated with moderate grasp pose quality $Q_p$, suggesting that imperfect pose estimation can contribute to these failures. However, clutter  $Q_c$ did not appear to be the main driver of CL, reinforcing the interpretation that fingertip flexibility is the critical factor. For collisions with surrounding objects (CLS), both Rigid and Fin-Ray grippers show relatively higher object scores $Q_o$, suggesting that these failures are more likely to occur with deformable objects. More importantly, CLS displayed the strongest dependency on clutter, consistently increasing under high $Q_c$. The relationship with pose quality $Q_p$ remains moderate, suggesting that CLS arises from a combination of spatial constraints and interaction dynamics rather than a single dominant predictor. Notably, CL and CLS were not observed for the suction gripper. 
\vspace{4pt}

In the rigid and Fin-Ray grippers, slip-related failures (SL) occurred across moderate–high $Q_c$ values, with a wide spread of clutter, indicating that slippage is not strongly driven by environmental density. The spread of object scores $Q_o$ suggests that slippage occurs across different object types, while moderate pose quality values $Q_p$ indicate that these failures are not primarily due to poor grasp pose estimation. Instead, SL is more likely influenced by contact conditions such as friction and force distribution. Grasp pose misalignment (WGP) failures across all three grippers were associated with higher $Q_o$ corresponding to deformable objects. This suggests that the 6D grasp pose generation models used in the system were not adequately trained for highly deformable objects, highlighting the need for such contributions in food waste sorting applications. However, WGP does not show a strong or consistent relationship with pose quality $Q_p$, suggesting that these failures are less dependent on the predicted grasp score and more influenced by the effectiveness of grasp pose generation and selection. Notably, WGP failures were not clutter-driven.
\vspace{4pt}

Finally, object non-conformance failures (ON), which are unique to the suction gripper, occur across a wide range of object scores $Q_o$ and are not confined to specific object categories. These failures are observed even at relatively high pose quality $Q_p$, indicating that they are not primarily caused by pose estimation errors. The influence of clutter remains moderate, suggesting that ON failures are mainly governed by surface characteristics and the ability to achieve a reliable seal rather than environmental complexity. 

\begin{table}[H]
\vspace{-6pt}
\caption{Impact of object-related (Object Score $Q_o$), pose quality-related  (Pose Score $Q_p$), and clutter-related (Clutter Score $Q_c$) pre-grasp conditions on the occurrence of failure modes across gripper types (corresponding to Fig.~\ref{boxplots}).}
\vspace{-6pt}
\renewcommand{\arraystretch}{1.3}
\centering
\begin{tabular}{l l c c c}
\toprule
&
\shortstack{\textbf{Failure}\\\textbf{Mode}} &
\shortstack{\textbf{$Q_o$}\\(Object Score)} &
\shortstack{\textbf{$Q_p$}\\(Pose Score)}&
\shortstack{\textbf{$Q_c$}\\(Clutter Score)} \\
\midrule
\multirow{4}{*}{\rotatebox{90}{\shortstack{Rigid /\\Fin-Ray}}}
& CL  & $\checkmark$ & $\triangle$ & $\times$ \\
& CLS & $\checkmark$ & $\triangle$ & $\checkmark$ \\
& SL  & $\triangle$ & $\triangle$ & $\triangle$ \\
& WGP & $\checkmark$ & $\triangle$ & $\times$ \\
\midrule
\multirow{2}{*}{\rotatebox{90}{Suction}}
& WGP & $\checkmark$ & $\triangle$ & $\times$ \\
& ON  & $\checkmark$ & $\times$ & $\times$ \\
\bottomrule
\end{tabular}

\begin{minipage}{0.95\linewidth}
\centering
\footnotesize
\vspace{4pt}
$\checkmark$: strong influence;\;
$\triangle$: moderate influence;\;
$\times$: weak or no influence.
\end{minipage}
\vspace{-22pt}
\label{failureinfluence}
\end{table}

\section{Conclusion}\label{sec6}

We presented GRAB, a comprehensive real-world benchmark for evaluating robotic grasping in cluttered food waste sorting scenarios. The primary strength of GRAB lies in its novel integration of 
object and environment-related pre-grasp conditions through graspability parameters, enabling a structured and interpretable analysis of how these factors influence grasp performance across different gripper modalities.  Unlike existing performance benchmarking approaches that rely primarily on success-based metrics or controlled datasets, GRAB enables performance evaluation under realistic, highly variable grasp-in-clutter (GIC) conditions using diverse deformable objects. 
\vspace{4pt}

Our investigation into the grasp performance of industrial grippers across seven contaminant categories, visualized via radar charts, reveals a clear performance dichotomy. The soft Fin-Ray gripper demonstrated superior resilience when handling deformable to semi-deformable objects, which constitute approximately 92\% of inorganic contaminants in food waste, consistently outperforming both the rigid and suction grippers. However, its performance was lowest on flat, thin-surfaced objects, which represent the remaining 8\% of contaminants. For these specific categories, the suction gripper achieved the highest overall performance, showing greater tolerance and effectiveness on rigid, planar surfaces.  This stark division in gripper efficacy indicates a crucial practical requirement: the integration of multiple modalities into a single, multi-modal gripper is necessary to achieve robust cross-category applicability in food waste sorting.
\vspace{4pt}

Statistical analysis further reveals a consistent hierarchy of influence among graspability parameters. Object quality emerges as the dominant factor governing grasp performance, acting as a strong positive predictor for Fin-Ray and rigid grippers while negatively influencing suction-based grasping. In contrast, pose quality and clutter exhibit comparatively moderate and context-dependent effects, suggesting that the advanced 6D perception pipeline significantly mitigates their influence.
\vspace{4pt}

From the failure mode analysis, it is noted that over 70\% of failures arise from physical interaction, including collisions with the work base or targeted object and slipping. This finding emphasizes the importance of incorporating structural flexibility and improved surface contact as key design factors for future gripper development for robust performance in food waste sorting. 
\vspace{4pt}

Overall, GRAB advances grasp-in-clutter benchmarking beyond performance evaluation by enabling interpretable analysis of pre-grasp conditions and their relationship to both success and failure within an advanced 6D grasping pipeline. This provides a principled foundation for designing adaptive and application-specific grasping systems and supports future research toward multimodal and failure-aware robotic manipulation in complex, unstructured environments. 
\vspace{24pt}

\medskip
\textbf{Acknowledgements} \par 
This work was jointly funded by a PhD scholarship awarded to Moniesha Thilakarathna from the University of Canberra and GOTERRA PTY LTD, Food Waste Sorting Company.

\vspace{-4pt}
\medskip
\textbf{Data Availability Statement} \par
The codes used and/or analysed during the current study are available at: \\
https://github.com/Moni9612/GRAB-Grasping-Real-World-Attribute-Benchmark

\vspace{-12pt}
\bibliographystyle{IEEEtran}  
\vspace{-4pt}
\bibliography{wileyNJD-Chicago}








\end{document}